\pdfoutput=1
\documentclass[11pt]{article}

\usepackage[preprint]{acl}
\usepackage{amsmath}

\newcommand\C[1]\null

\usepackage{times}
\usepackage{latexsym}
\usepackage{amsfonts}
\usepackage{algorithm}
\usepackage{algorithmic}

\usepackage[T1]{fontenc}
\usepackage{dblfloatfix}

\usepackage{listings}
\usepackage{xcolor}

\usepackage[utf8]{inputenc}

\usepackage{microtype}

\usepackage{inconsolata}

\usepackage{graphicx}

\usepackage{float}
\title{SCODEL: Shared Canonical Ontology Definitions for Food Entity Linking and Ontology-Grounded Recipe Knowledge Graphs}

\author{
\begin{tabular}{c}
Nuhu Ibrahim, Rishi Ravikumar, Ayuba Dauda, Fuqi Xu,\\
Robert Stevens, Riza Batista-Navarro
\end{tabular}
\\[0.5em]
Department of Computer Science, The University of Manchester, United Kingdom
\\
\textbf{Correspondence:} 
\texttt{nuhu.ibrahim@manchester.ac.uk}
}

\usepackage{multirow}

\begin{document}
\maketitle
\begin{abstract}
Identifying recipe entities and linking them to ontology concepts is fundamental for semantic recipe understanding. However, food mentions in recipes are often ambiguous, context-dependent, and lexically diverse, making ontology entity linking (EL) challenging. We propose SCODEL, a streamlined, model-agnostic approach based on a central idea: contextual food mentions and ontology concepts that refer to the same entity should learn the same canonical ontology-oriented definition. SCODEL uses these shared definitions as a semantic normalisation layer, providing complementary semantic information to existing EL models without requiring a new linking architecture. We further introduce ReciLinkD, the first benchmark for contextual food ontology entity linking in recipes. Experiments show that SCODEL consistently improves three modular EL models (ReFinED, GSM-EL, and BLINK): SCODEL$\times$BLINK achieves the highest Hit@1 on seen concepts (88.33\%), while SCODEL$\times$GSM-EL achieves the highest Hit@1 on unseen concepts (83.15\%), outperforming the strongest large language model (LLM) based EL baseline in each setting. Experiments on a biomedical ontology benchmark further demonstrate that SCODEL generalises beyond the food domain. Finally, we use SCODEL to ground food mentions at scale and construct an ontology-grounded recipe knowledge graph, demonstrating its utility for ontology-aware recipe retrieval and corpus-level food analysis.
\end{abstract}

\section{Introduction}
Entity linking (EL) is a fundamental task in natural language processing (NLP) that grounds textual mentions to canonical concepts in structured knowledge sources such as ontologies and knowledge graphs. By linking surface forms to structured concepts, EL enriches textual data with formal semantics, enabling interoperability, semantic reasoning, and knowledge integration across heterogeneous datasets \cite{rao2012entity,shen2014entity}. Consequently, EL has become an essential component of biomedical and scientific NLP, supporting downstream applications including information extraction, knowledge graph construction, and semantic search.

Despite its importance, ontology entity linking remains relatively underexplored in the food domain. Food mentions may differ substantially from ontology labels despite referring to the same food concept (e.g., \textit{garbanzo beans} vs.\ \textit{chickpea}), require recipe context to resolve underspecified mentions (e.g., \textit{stock} referring to \textit{chicken stock} or \textit{vegetable stock}), or occur at different levels of specificity in the ontology hierarchy (e.g., \textit{chilli pepper} vs.\ \textit{jalapeno pepper}). Existing approaches are typically limited by small annotated datasets, restricted ontology coverage, or memorisation-based methods that struggle to generalise to previously unseen concepts \cite{pappu2017lightweight,chong2018implicit,popovski2019foodontomap,feher2023few,gjorgjevikj2025foodsem}. 

At the same time, complementary resources have established a rich ecosystem for computational food research. FoodOn \cite{dooley2018foodon} provides a structured ontology of food concepts, while Recipe1M \cite{marin2018recipe1m+} and RecipeNLG \cite{bien2020recipenlg} provide large-scale recipe corpora. ReciFine \cite{nuhuibrahim2024recifine} further provides fine-grained food entity annotations over recipe text. However, the absence of robust contextual entity linking prevents effective integration between these resources, limiting semantic recipe understanding and downstream analyses such as nutritional reasoning, dietary inference, ingredient composition analysis, and food analytics through large-scale recipe knowledge graph construction.

Recent advances in entity linking have improved zero-shot generalisation through dense representations and pretrained language models \cite{moreno2017combining,hou2020improving,parravicini2019fast,logeswaran2019zero,yao2020zero,wu2020scalable,ayoola2022refined}. Nevertheless, these approaches primarily operate on lexical mentions or contextual embeddings, leaving a persistent semantic gap between informal recipe language and structured ontology concepts. This challenge is especially pronounced in recipes, where ingredient mentions vary widely and depend heavily on cooking context.

To address these challenges, we propose SCODEL, a method that learns shared canonical ontology-oriented definitions for both contextual recipe food mentions and ontology concepts, projecting them into a shared semantic definition space before downstream entity linking. Rather than replacing existing entity linking systems, SCODEL serves as a model-agnostic semantic normalisation layer that can be integrated with modular entity linking methods to improve supervised and zero-shot linking performance. We further introduce ReciLinkD, the first benchmark for contextual recipe ontology entity linking, covering food entities as they appear in recipes. Together, SCODEL and ReciLinkD enable recipe-level ontology grounding, facilitating richer semantic understanding and analysis of large-scale recipe collections. All code, datasets, and trained model artefacts associated with this work are released publicly under CC BY-NC 4.0\footnote{\href{https://github.com/nuhu-ibrahim/ReciLinkD}{https://github.com/nuhu-ibrahim/ReciLinkD}}.

Our contributions are summarised as follows:
\begin{itemize}
    \item We propose SCODEL, a streamlined, model-agnostic approach that learns shared canonical ontology-oriented definitions as a semantic normalisation layer for ontology entity linking.

    \item We introduce ReciLinkD, the first contextual recipe ontology entity-linking benchmark, with 3,979 manually annotated pairs, expanded to 15{,}559 manually verified pairs using SCODEL.

    \item We demonstrate that SCODEL consistently improves diverse modular entity linking frameworks across both supervised and zero-shot settings.

    \item We provide extensive ablation studies analysing SCODEL's key design choices, demonstrate its generalisability beyond food ontology on the GSCplus dataset with the Human Phenotype Ontology~\citep{luo2021phenotagger}, and showcase the practical applicability of food entity linking through a downstream recipe knowledge graph construction application.
\end{itemize}

\section{Related Work}
\subsection{Entity Linking}
Early EL approaches relied on lexical matching and handcrafted features \cite{rao2012entity,shen2014entity}, while more recent methods employ dense neural representations, graph reasoning, and pretrained language models to improve semantic matching and candidate ranking \cite{parravicini2019fast,hou2020improving}. Modern EL frameworks, including ReFinED \cite{ayoola2022refined}, BLINK \cite{laskar2022blink}, and GSM-EL \cite{cheng2024gsm}, have demonstrated strong performance through scalable retrieval and ranking architectures, enabling effective supervised and zero-shot entity linking.

LLMs have recently been incorporated into EL pipelines to further improve semantic understanding. LLMAEL \cite{xin2025llmael} augments mention representations by prompting an LLM to generate additional contextual information before candidate ranking, while LLM4EL \cite{xu2026llm4el} generates semantic descriptions for both food mentions and ontology concepts to facilitate ontology-aware matching in zero-shot settings. Similarly, \citet{vollmers2025contextual} use LLM-generated contextual augmentation to improve entity disambiguation. SCODEL is most closely related to LLM4EL, as both approaches generate definitions for the food mention and ontology concept sides. However, LLM4EL generates food mention and ontology concept definitions separately through zero-shot prompting, without learning a common definition target across the two views, and performs entity linking through embedding-based semantic similarity between the generated definitions. In contrast, SCODEL learns shared canonical definition generation from aligned food mention and ontology concept pairs: the contextual food mention and the metadata of its corresponding ontology concept are independently learned to generate a similar canonical ontology-oriented definition. Thus, the distinction is not the use of generated definitions itself, but the explicit learning of a shared canonical target that semantically aligns the two views before entity linking, as demonstrated in Appendix~\ref{app:semantic_alignment_analysis}. Unlike LLM4EL's direct similarity-based linking, SCODEL integrates the learned canonicalisation with diverse downstream EL models. Furthermore, while LLM4EL was developed for biomedical entity linking, SCODEL extends this definition-based approach to the challenging domain of food ontology entity linking.

\subsection{Recipe and Food Entity Linking}
Compared with biomedical and general-domain EL, food ontology entity linking remains relatively underexplored. Existing work has focused on ontology alignment, food concept mapping, or structured food knowledge extraction \cite{pappu2017lightweight,chong2018implicit,popovski2019foodontomap, ibrahim2026knowledge}. More recent studies have explored food EL using small annotated datasets or LLM-based approaches \cite{feher2023few,gjorgjevikj2025foodsem}. However, these methods are generally limited to conventional food names, narrow ontology coverage, or memorisation-based linking, and rarely consider context-dependent recipe food mentions or generalisation to previously unseen ontology concepts. 

\begin{figure*}[t]
    \centering
    \includegraphics[
        width=0.80\textwidth,
        trim=0 10cm 0 0,
        clip
    ]{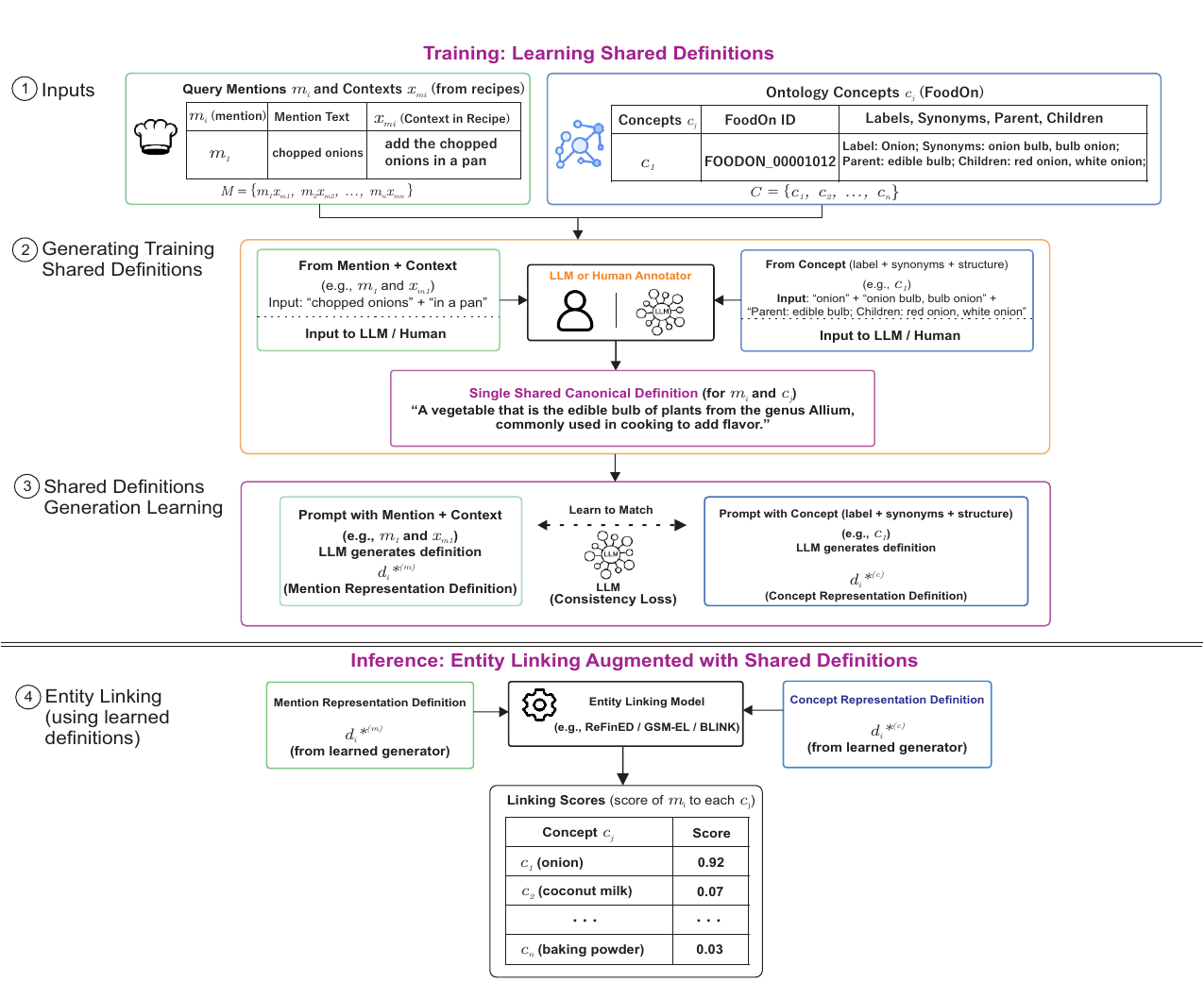}
    \caption{Overview of the proposed SCODEL methodology.}
    \label{fig:scodel-architecture}
\end{figure*}

\section{Food EL Task Formulation}
\label{subsec:task-formulation}
Given a recipe food mention $m$ and its surrounding context $x_m$, where $x_m$ denotes the recipe context in which $m$ occurs, the entity linking task is to identify the corresponding concept in a structured food ontology. Let $\mathcal{C}_{\mathrm{Food}}$ denote the set of food concepts in the ontology. An entity linking model assigns a compatibility score

{\small 
\[
s(c \mid m,x_m)
\]
}

to each candidate concept $c \in \mathcal{C}_{\mathrm{Food}}$. The predicted concept is

{\small 
\[
\hat{c}
=
\operatorname*{arg\,max}_{c \in \mathcal{C}_{\mathrm{Food}}}
s(c \mid m,x_m).
\]
}

\paragraph{Zero-shot Entity Linking}
Let $\mathcal{C}^{\mathrm{train}}_{\mathrm{Food}}$ denote the FoodOn concepts represented in the training data. We define a test instance as zero-shot when its gold ontology concept $c$ is completely held out from training. Specifically, neither the food mentions and ontology concepts associated with $c$ nor their canonical definitions are used to train the SCODEL definition generator or the downstream entity linking model:

{\small 
\[
c \notin \mathcal{C}^{\mathrm{train}}_{\mathrm{Food}}.
\]
}

At inference, the held-out ontology concept and its metadata remain available in the candidate ontology for matching. Thus, zero-shot evaluation assesses whether SCODEL and the downstream entity linker can generalise to unseen ontology concepts without having received concept-specific supervision during either shared canonical definition learning or downstream entity linker training.

\section{Methodology}
\label{sec:methodology}
We propose SCODEL, a semantic enhancement method for ontology entity linking. Rather than relying only on the original food mentions and ontology concepts, SCODEL augments both representations with ontology-oriented definitions generated by a language model trained to produce shared canonical definitions. These definitions provide complementary semantic signal that reduces the gap between informal recipe language and structured ontology descriptions while remaining fully compatible with existing entity linking architectures.

As illustrated in Figure~\ref{fig:scodel-architecture}, SCODEL consists of three stages: shared canonical definition learning, canonical definition generation, and entity linking integration.

\subsection{Shared Canonical Definition Learning}
Recipe food mentions and ontology concepts describe the same culinary concepts using fundamentally different linguistic styles. Recipe food mentions are short contextual expressions extracted from recipes, e.g., \textit{parm} used in the context \textit{sprinkle freshly grated parm over the pasta before serving}, whereas ontology concepts are represented using structured lexical and taxonomic information, e.g., \textit{parmesan cheese (FOODON\_03303675)}. Directly comparing these heterogeneous representations often results in semantic mismatch. To bridge this gap, SCODEL learns to generate a shared canonical ontology-oriented definition that provides a unified semantic description for both representations.

To supervise this process, we manually construct a collection of canonical ontology-oriented definitions. For every linked recipe food mention and ontology concept pair, annotators produce a single ontology-oriented definition describing the underlying culinary concept. Formally, for the $i$th aligned pair, the shared canonical definition is

{\scriptsize 
\[
d_i^{*}
=
\mathcal{A}
\left(
m_i,
x_{m_i},
c_i^{\mathrm{meta}}
\right),
\]
}

where $\mathcal{A}$ denotes the manual annotation process, $m_i$ is the recipe food mention, $x_{m_i}$ is its surrounding context, and $c_i^{\mathrm{meta}}$ is the structured representation of its linked ontology concept. The concept representation is constructed as

{\scriptsize 
\[
\begin{aligned}
c_i^{\mathrm{meta}}
=
\operatorname{concat}(
&\operatorname{label}(c_i),
\operatorname{synonyms}(c_i),\\
&\operatorname{parents}(c_i),
\operatorname{children}(c_i)).
\end{aligned}
\]
}

Thus, $d_i^{*}$ is not specific to either the recipe or ontology representation; rather, it is a single canonical definition grounded jointly in both and serves as their shared supervision target.

SCODEL trains a shared definition generation language model $G_{\theta}$ to generate the canonical definition independently from either representation:

{\scriptsize 
\[
G_{\theta}(m_i,x_{m_i}) \rightarrow d_i^{*},
\qquad
G_{\theta}(c_i^{\mathrm{meta}}) \rightarrow d_i^{*}.
\]
}

$G_{\theta}$ is jointly optimised over a training set of $N$ aligned manually annotated food mention and ontology concept pairs using the learning objective

{\scriptsize 
\[
\begin{aligned}
\mathcal{L}
=
-\sum_{i=1}^{N}
\Big(
&
\log P_{\theta}
(d_i^{*}\mid m_i,x_{m_i})
+  
\log P_{\theta}
(d_i^{*}\mid c_i^{\mathrm{meta}})
\Big).
\end{aligned}
\]}

By optimising both food mentions and ontology concepts towards the same manually curated target, SCODEL learns to produce semantically consistent, ontology-oriented definitions from the two heterogeneous input representations.

\subsection{Canonical Definition Generation}
During inference, the shared language model $G_\theta$ generates canonical ontology-oriented definitions for recipe food mentions and ontology concepts. Since definitions are generated directly from recipe food mentions or ontology concept metadata, SCODEL naturally supports previously unseen recipe food mentions or ontology concepts without requiring additional manual annotation.

\subsection{Entity Linking Integration}
SCODEL is model-agnostic and does not replace existing entity linking architectures. Instead, the generated canonical definitions are incorporated as complementary semantic representations alongside the original entity linking inputs.

For recipe food mentions, the entity linking model receives the original food mention and its surrounding context, augmented with the generated canonical definition. Similarly, each ontology concept is represented by its original ontology metadata and augmented with the corresponding generated canonical definition.

The augmented representations are then supplied to existing entity linking models for candidate retrieval and ranking without modifying their underlying architectures. 

\section{Our ReciLinkD Dataset}
\label{sec:recilinked-dataset}
We introduce ReciLinkD, a gold-standard food entity linking dataset aligning 3{,}979 recipe-derived food mentions with FoodOn ontology concepts. 

\subsection{Manual Annotation Process}
We perform manual entity linking. For each selected recipe-sourced food mention, annotators are presented with its surface form and representative contextual examples from recipes. Annotators are then asked to select the most semantically appropriate FoodOn concept. The annotation workload is distributed across four annotators with domain familiarity. The annotation guide is provided in Appendix~\ref{appendix:manual_recipe_mention_linking_guideline}.

To assess annotation reliability, 300 randomly selected food mentions were independently annotated by all four annotators. Across all six annotator pairs, the mean pairwise F1-score was 89.67\%, and the mean Cohen's $\kappa$ was 0.7253, indicating strong agreement. The full pairwise results are in Appendix~\ref{appendix:annotation_agreement}.

\subsection{ReciLinkD Dataset Descriptive Metrics}
\label{appendix:dataset_metrics}
\paragraph{FoodOn Ontology:}
\label{sec:ontology_extraction}
FoodOn \cite{dooley2018foodon} is a structured ontology covering a broad range of food-related concepts, including food materials, food processing and preparation, biological source organisms, packaging, and geographic locations. As our entity linking task focuses specifically on food mentions in recipes, we restrict the candidate ontology space to food material concepts. After normalisation and deduplication, this yields 1{,}014 food-related concepts. The extraction procedure is described in Appendix~\ref{appendix:foodon_extract}.
\paragraph{ReciFine Dataset:}
ReciFine \cite{nuhuibrahim2024recifine} provides fine-grained food annotations extracted from approximately 2.2M recipes using methods described in \citet{ibrahim2026knowledge}. From this corpus, we select the 4{,}500 most frequent food mentions and the contexts in which they appear in recipes, forming the recipe-sourced food mention set for manual linking. Frequency-based selection ensures broad coverage of commonly occurring culinary concepts.

After manual annotation, 3{,}979 of the 4{,}500 selected food mentions were successfully mapped to 979 FoodOn concepts. Unlinked food mentions primarily consisted of context-dependent referring expressions, such as \textit{it} in the context \textit{mix the flour, butter, and sugar until a crumbly mixture forms, then spread it over the fruit}, where the referent is identifiable from the full recipe context to be a \textit{dough} but the food mention itself, i.e., \textit{it}, does not explicitly name the food entity. These were excluded to maintain a gold standard centred on explicit food mentions. For the remaining 35 of the 1{,}014 candidate FoodOn concepts, no suitable food mention was identified among the selected ReciFine mentions; these concepts are therefore not represented in ReciLinkD. Further details on the ReciLinkD dataset statistics are provided in Appendix~\ref{appendix:foodon_extract} (Table~\ref{tab:food_stats}).

\subsection{Manual Canonical Definition Generation}
\label{sec:def_generation}
To construct training data for SCODEL, two annotators manually created a single ontology-oriented canonical definition for each aligned recipe food mention and ontology concept pair.

To establish a consistent annotation procedure, both annotators independently generated canonical definitions for 50 randomly selected aligned pairs. They compared their definitions for consistency in semantic interpretation, terminology, and level of abstraction and detail, rather than exact textual agreement. Substantive disagreements were discussed with an independent assessor and resolved by consensus, with the resulting conventions incorporated into the annotation guidelines. The remaining pairs were shared between the annotators and annotated using the agreed guidelines. Full guidelines are provided in Appendix~\ref{appendix:manual_definition_guideline}.

\section{Experimental Setup}
\subsection{Canonical Definition Generation Models}
We evaluate three open-weight LLMs as backbone models for learning and generating shared canonical definitions: Llama-3-8B, Mistral-7B, and Qwen2.5-7B. Each model is trained on the manually aligned canonical definitions described in Section~\ref{sec:def_generation} and subsequently generates canonical definitions independently.

\begin{table*}[ht!]
\centering
\scriptsize
\setlength{\tabcolsep}{5pt}
\begin{tabular}{|l|cccc|cccc|c|}
\hline

\multirow{2}{*}{\textbf{Method}}
&
\multicolumn{4}{c|}{\textbf{Supervised (Seen)}}
&
\multicolumn{4}{c|}{\textbf{Zero-Shot (Unseen)}}
&
\textbf{AVG} \\
\cline{2-10}

&
\textbf{Hit@1}
&
\textbf{Hit@5}
&
\textbf{MRR}
&
\textbf{H-Acc}
&
\textbf{Hit@1}
&
\textbf{Hit@5}
&
\textbf{MRR}
&
\textbf{H-Acc}
&
\textbf{Hit@1}
\\
\hline

LLM4EL
& 63.29 & 88.33 & 73.90 & 66.89
& 63.72 & 90.21 & 74.92 & 71.05
& 63.51 \\

LLMAEL
& 85.25 & \underline{94.51} & 89.61 & \underline{86.28}
& 80.71 & \underline{94.70} & 86.47 & 81.52
& \underline{82.98} \\

LLM Rerank@10
& 82.68 & 89.88 & 85.97 & 83.70
& 83.02 & 92.80 & 87.19 & 83.83
& 82.85 \\

LLM Rerank@50
& 81.13 & 88.51 & 84.52 & 81.65
& 80.43 & 90.35 & 84.87 & 81.39
& 80.78 \\
\hline

Lexical Matching
& 52.66 & 73.93 & 61.62 & 54.15
& 56.79 & 75.27 & 65.81 & 57.99
& 54.73 \\

ReFinED
& 81.82 & 89.71 & 85.42 & 82.68
& 78.40 & 92.66 & 84.38 & 79.62
& 80.11 \\

GSM-EL
& 76.84 & 90.56 & 82.62 & 78.38
& \underline{82.33} & 93.75 & \underline{87.48} & \underline{84.78}
& 79.59 \\

BLINK
& \underline{87.47} & 93.65 & \underline{90.31} & \textbf{88.85}
& 71.19 & 87.09 & 77.90 & 74.72
& 79.33 \\
\hline

SCODEL $\times$ ReFinED
& 85.08 & 93.31 & 89.11 & 85.59
& 82.07 & 93.75 & 87.38 & 83.02
& \textbf{83.58} \\

SCODEL $\times$ GSM-EL
& 80.78 & 92.62 & 86.24 & 82.67
& \textbf{83.15} & \textbf{94.83} & \textbf{88.38} & \textbf{85.59}
& 81.97 \\

SCODEL $\times$ BLINK
& \textbf{88.33} & \textbf{94.68} & \textbf{90.95} & \textbf{88.85}
& 75.00 & 88.72 & 81.32 & 79.21
& 81.67 \\
\hline

\end{tabular}
\caption{Main entity linking results on the ReciLinkD benchmark under seen and unseen settings. SCODEL uses manually generated canonical definitions as supervision and Mistral-7B, the best-performing individual definition-learning backbone (Table~\ref{tab:backbone_ablation}). LLM4EL uses Mistral-7B for both definition generation and embedding generation, LLMAEL uses Mistral-7B for context augmentation and ReFinED for entity linking (Appendix~\ref{app:llmael_other_el_models}: Table~\ref{tab:llmael_el_models}), and LLM Rerank@10 and Rerank@50 use Mistral-7B for reranking the ReFinED predictions. Best results are bold, second-best results are underlined, and AVG denotes the mean Hit@1 across both settings.}
\label{tab:main_results}
\end{table*}

\subsection{Entity Linking Models}
SCODEL is model-agnostic and is evaluated with three SOTA entity linking models: ReFinED, GSM-EL, and BLINK. Because these models were originally developed for general-domain entity linking, primarily using encyclopedic knowledge bases such as Wikipedia and Wikidata rather than specialised food ontologies such as FoodOn, we made minor adaptations where necessary. Appendix~\ref{appendix:el_model_adaptations} provides further details on the modifications made to each modular entity linking model.

\subsection{Baselines}
We compare SCODEL against two categories of baselines: (1) LLM-based entity linking frameworks, including LLMAEL, LLM4EL and LLM Rerank, that augment entity linking using LLM-generated semantic representations; and (2) standalone entity linking models, namely ReFinED, GSM-EL, and BLINK, evaluated without SCODEL augmentation. Lexical matching is also evaluated as a simple non-neural baseline. Appendix~\ref {appendix:model_descriptions} and Appendix~\ref{appendix:el_model_adaptations} provide further details on the generic EL models and their implementation settings, and describe modular EL model-specific adaptations. The LLM re-ranking and lexical matching procedures are detailed in Appendices~\ref{appendix:llm-re-ranking-entity-linking} and~\ref{appendix:lexical-matching}, respectively, while dataset splits are described in Appendix~\ref{appendix:data_splits}.

\subsection{Evaluation}
We report Hit@1, Hit@5, Mean Reciprocal Rank (MRR), and hierarchy-aware accuracy (H-Acc). Appendix~\ref{appendix:evaluation_metrics} provides detailed definitions of all evaluation metrics.

\section{Results and Discussion}
Table~\ref{tab:main_results} presents the main entity linking results. Across all modular entity linking models, SCODEL consistently improves performance over their original implementations, demonstrating that shared canonical definition learning is complementary to existing EL architectures. SCODEL achieves the best overall average performance when combined with ReFinED (83.58\% Hit@1), while SCODEL$\times$BLINK attains the highest supervised performance (88.33\% Hit@1, 90.95\% MRR). More importantly, SCODEL$\times$GSM-EL achieves the strongest zero-shot results (83.15\% Hit@1, 88.38\% MRR). LLM4EL performs comparatively lower than the other EL methods, which may reflect differences in their linking mechanisms: LLM4EL directly matches generated mentions and concept definitions through embedding similarity, whereas the other methods use trained entity-linking and ranking architectures. This suggests that, on ReciLinkD, definition-based representations are more effective when integrated with a trained EL model than when used for direct similarity-based matching. The slightly higher zero-shot performance of some models may partly reflect lexical differences between the splits, as exact food mention and ontology concept label matches are more frequent in the zero-shot set than in the supervised set (18.61\% vs.\ 11.84\%; Appendix~\ref{appendix:data_splits}: Table~\ref{tab:split_stats}). Overall, these results demonstrate that learning shared canonical definitions improves semantic alignment between recipe food mentions and ontology concepts, supporting generalisation to unseen concepts while remaining complementary to different entity linking architectures.

\subsection{Ablation Studies}
We conduct ablation studies to examine the key design choices underlying SCODEL and to determine where its performance gains originate. We focus on the contribution of shared canonical definition learning and the effect of different sources of canonical supervision, and provide additional analyses of model configuration, semantic alignment, and other experimental choices in Appendix~\ref{appendix:further_ablation_study}.

\subsubsection{SCODEL Cross-Domain Generalisation} 
To assess generalisability, we apply SCODEL without modification to the GSCplus biomedical entity linking benchmark. As shown in Table~\ref{tab:gsc_transfer}, SCODEL consistently improves ReFinED and achieves the best overall performance, demonstrating that the proposed shared canonical definition learning method generalises beyond food ontology entity linking. The shared canonical definitions for GSCplus are learned and generated using Llama-3-8B to maintain consistency with the backbone model used by \cite{xu2026llm4el}.

\begin{table}[ht!]
\centering
\scriptsize
\setlength{\tabcolsep}{6pt}
\begin{tabular}{|l|c|c|c|}
\hline
\textbf{Method} & \textbf{Hit@1} & \textbf{Hit@5} & \textbf{MRR} \\
\hline
ReFinED Baseline & 62.75 & 77.36 & 68.96 \\
\hline
LLM4EL & 63.00 & 81.10  & -- \\
LLMAEL & \underline{69.97} & \underline{85.73}  & \underline{77.08} \\
SCODEL & \textbf{72.45} & \textbf{87.87}  & \textbf{79.35} \\
\hline
\end{tabular}
\caption{Results on the GSCplus dataset with the Human Phenotype Ontology~\citep{luo2021phenotagger} using ReFinED as the downstream entity linking model for LLMAEL and SCODEL. LLM4EL results are taken directly from \citet{xu2026llm4el}, which does not report MRR; all other results are obtained using our implementation on the same benchmark. Best results are bold and second-best results are underlined.} 
\label{tab:gsc_transfer}
\end{table}

\subsubsection{Effect of Shared Definition Learning}
Table~\ref{tab:shared_definition_ablation} evaluates the contribution of shared definition learning. \textit{Mention Definition Learning Only} trains the definition generator using definitions of recipe food mentions and their contexts alone; \textit{Ontology Definition Learning Only} uses definitions of ontology metadata alone, while SCODEL jointly learns a shared canonical definition space from both representations. Although \textit{Ontology Definition Learning Only} also improves entity linking over the ReFinED baseline, SCODEL achieves the best performance, demonstrating that jointly learning from recipe and ontology representations produces more effective semantic definitions for downstream entity linking.

\begin{table}[ht!]
\centering
\scriptsize
\setlength{\tabcolsep}{8pt}
\begin{tabular}{|l|c|c|}
\hline
\textbf{Training Strategy} &
\textbf{Hit@1} &
\textbf{MRR} \\
\hline

ReFinED Baseline  & 81.82 & 85.42 \\
\hline
Mention Def. Learning Only  & 81.65 & 86.85 \\
Ontology Def. Learning Only & \underline{82.68} & \underline{86.68} \\
SCODEL                           & \textbf{85.08} & \textbf{89.11} \\
\hline
\end{tabular}
\caption{Effect of different definition learning strategies using ReFinED as the downstream entity linking model. Best results are bold and second-best results are underlined.}
\label{tab:shared_definition_ablation}
\end{table}

\subsubsection{Frozen Definition Generators} 
Table~\ref{tab:frozen_generators} compares SCODEL with frozen definition generators that did not learn food mentions and ontology concepts shared definitions. While frozen LLM-generated definitions provide modest improvements over the ReFinED baseline, SCODEL achieves higher Hit@1 and MRR, showing that learning shared canonical definitions is more effective than using off-the-shelf LLM-generated definitions without adaptation.

\begin{table}[ht!]
\centering
\scriptsize
\setlength{\tabcolsep}{8pt}
\begin{tabular}{|l|c|c|}
\hline
\textbf{Definition Generator} &
\textbf{Hit@1} &
\textbf{MRR} \\
\hline
ReFinED Baseline  & 81.82 & 85.42 \\
\hline
Frozen Mistral-7B & 82.16 & \underline{87.46} \\
Frozen GPT-5.1-2025-11-13 & \underline{82.50} & 86.93 \\
\hline
SCODEL (Mistral-7B) & \textbf{85.08} & \textbf{89.11} \\
\hline
\end{tabular}
\caption{Comparison of SCODEL with frozen definition generators that did not learn shared definitions. ReFinED is used as the downstream EL model. Best results are bold and second-best results are underlined.}
\label{tab:frozen_generators}
\end{table}

\subsubsection{Definition Learning Backbone}
Table~\ref{tab:backbone_ablation} compares comparable-size backbone LLMs (7--8B parameters) for SCODEL. Hard voting selects the most predicted ontology concept, whereas soft voting selects the ontology concept with the highest aggregated confidence score. Among the individual models, Mistral-7B achieves the highest Hit@1 and is therefore selected as the definition-learning backbone for the main experiments. Although ensembling further improves performance by combining predictions from multiple definition-learning backbones, we use a single Mistral-7B backbone in the main experiments to maintain a consistent single-model configuration across downstream EL models. All model sizes are trained using the same SCODEL training setup.

\begin{table}[ht!]
\centering
\scriptsize
\setlength{\tabcolsep}{8pt}
\begin{tabular}{|l|l|c|c|}
\hline
\textbf{Method} &
\textbf{Def. Generator} &
\textbf{Hit@1} &
\textbf{MRR} \\
\hline
\multirow{3}{*}{Single}
& Llama-3-8B & 82.33 & 86.43 \\
& Mistral-7B & 85.08 & 89.11 \\
& Qwen2.5-7B & 83.88 & 87.94 \\
\hline

\multirow{2}{*}{Ensemble}
& Hard Voting & 85.42 & 90.16 \\
& Soft Voting & 86.11 & 90.59 \\
\hline

\end{tabular}
\caption{Comparison of backbone models for SCODEL shared definition learning and generation. ReFinED is used as the downstream entity linking model.}
\label{tab:backbone_ablation}
\end{table}

\subsection{Error Analysis}
Error analysis of the SCODEL entity linker (ReFinED) shows that most mistakes arise from confusion between semantically related concepts at different levels of the ontology hierarchy (Table~\ref{tab:qualitative_examples}). Less frequently, the model predicts conceptually related but incorrect concepts, such as \textit{sweet cherry} instead of \textit{tomato}. These errors suggest that SCODEL captures the underlying semantics well but occasionally struggles to distinguish the appropriate level of abstraction.

\begin{table*}[ht!]
\scriptsize
\centering
\setlength{\tabcolsep}{5pt}
\begin{tabular}{|p{5.2cm}|l|l|}
\hline
\textbf{Mention with Context} &
\textbf{Gold Concept} &
\textbf{Predicted Concept} \\
\hline

Add the \textit{dairy milk} to the sauce and stir until smooth.
&
animal milk (\href{http://purl.obolibrary.org/obo/FOODON_02010013}{FOODON\_02010013})
&
cow milk (\href{http://purl.obolibrary.org/obo/FOODON_03307455}{FOODON\_03307455}) \\
\hline

Toss the salad with fresh \textit{snap peas} before serving.
&
pea (\href{http://purl.obolibrary.org/obo/FOODON_03301300}{FOODON\_03301300})
&
sugar snap pea plant (\href{http://purl.obolibrary.org/obo/FOODON_03412467}{FOODON\_03412467}) \\
\hline

Garnish with sliced \textit{cherry tomatoes} before serving.
&
tomato (\href{http://purl.obolibrary.org/obo/FOODON_03301453}{FOODON\_03301453})
&
sweet cherry (\href{http://purl.obolibrary.org/obo/FOODON_03311457}{FOODON\_03311457}) \\
\hline

Roast the \textit{baby potatoes} until golden and tender.
&
potato (\href{http://purl.obolibrary.org/obo/FOODON_03301449}{FOODON\_03301449})
&
red bliss potato (\href{http://purl.obolibrary.org/obo/FOODON_03307121}{FOODON\_03307121}) \\
\hline

\end{tabular}
\caption{Examples of common entity linking errors made by the SCODEL (Mistral-7B) x ReFinED model.}
\label{tab:qualitative_examples}
\end{table*}

\section{Knowledge Graph Applications}
\subsection{Scaling Food Entity Linking to ReciFine}
To evaluate SCODEL beyond ReciLinkD, we apply SCODEL $\times$ ReFinED, the most consistent configuration across seen and unseen concepts, to 20{,}000 food mentions from the larger ReciFine corpus and manually evaluate the predicted FoodOn links. Overall, 57.90\% of the links were judged correct and suitable for use as gold-standard pairs, 36.13\% were judged correct but unsuitable as standalone food mentions, and 10.97\% were judged incorrect. This provides a large-scale assessment of generalisation beyond ReciLinkD while substantially expanding our manually verified food mention and ontology concept pairs from 3{,}979 to 15{,}559. Appendix~\ref{appendix:recifine_linking} provides further details on food mention selection, manual evaluation, and error categories.

\subsection{Ontology-Grounded Recipe Knowledge Graph}
Using the 3{,}979 gold-standard links from ReciLinkD and the new 11{,}580 manually verified and automatically linked pairs, we extend ontology grounding across the 2.2M recipes in ReciFine by connecting their food mentions to FoodOn concepts and their ontology hierarchy. The resulting knowledge graph contains over 2M grounded recipes linked to at least one FoodOn concept, 19{,}504 grounded lexically unique food mentions, and 3{,}188 direct and inherited grounded FoodOn concepts, enabling recipes to be queried and analysed through shared ontology concepts and higher-level semantic categories rather than surface forms alone. Appendix~\ref {appendix:kg_details} provides further details on graph construction and statistics.

\subsection{Ontology-Aware Recipe Retrieval and Corpus Analysis}
Across 15 crafted food-category queries, ontology-aware retrieval substantially outperforms lexical and semantic retrieval, improving macro-F1 by 25.42 percentage points over semantic retrieval and 33.76 percentage points over lexical retrieval (Appendix~\ref{appendix:ontology_aware_recipe_retrieval}: Table~\ref{tab:ontology_retrieval}). The advantage is particularly pronounced for conceptually broad queries: for example, \textit{winter squash} achieves 74.51\% F1 with ontology-aware retrieval, compared with 0.00\% and 20.83\% for lexical and semantic retrieval, respectively, allowing recipes containing more specific descendant foods to be recovered without explicitly mentioning the queried category. Importantly, these results are obtained while fewer than 10\% of ReciFine food mentions are currently grounded in the graph, suggesting substantial scope for increased coverage as further food mentions are linked. Beyond retrieval, ontology propagation transforms individual food mention-ontology concept links into corpus-level semantic evidence across recipes (Appendix~\ref{appendix:food_category_prevalence}: Table~\ref{tab:category_prevalence}). It also enables quantification of relationships between food categories: for example, 74.74\% of recipes containing plant food products also contain animal food products (1.3M recipes), while 68.69\% of recipes containing vegetables also contain fruit (Appendix~\ref{appendix:food_category_cooccurence}: Table~\ref{tab:category_cooccurrence}). Appendix~\ref{appendix:kg_analysis} provides further results, qualitative retrieval examples, and details of the human evaluation and LLM-as-a-judge methodologies used to assess the retrieval results and corpus-level analysis.

\section{Conclusion}
In this work, we introduced SCODEL, a streamlined, model-agnostic method for ontology entity linking that learns a shared canonical definition space for contextual recipe food mentions and ontology concepts before downstream entity linking. By jointly learning shared, ontology-oriented definitions from both representations, SCODEL provides a semantic normalisation layer that moves beyond surface-form matching and better handles ambiguous, context-dependent food mentions. Extensive experiments show consistent improvements over strong modular entity linking baselines in both supervised and zero-shot settings, and demonstrate that the method generalises beyond the food domain to biomedical ontologies. Furthermore, by enabling robust integration of recipe corpora with FoodOn, SCODEL facilitates downstream applications such as ontology-grounded recipe knowledge graph construction and semantic food analysis. More broadly, the proposed method provides a general and modular approach to ontology entity linking in domains where continual growth of ontology concepts presents significant challenges.

\section*{Limitations}
SCODEL is designed to exploit the semantic information available in structured ontologies, and its effectiveness can therefore depend on the quality and completeness of ontology metadata. FoodOn provides rich information including concept labels, definitions, synonyms, and hierarchical relations, whereas ontologies with substantially sparser metadata may provide less information from which to generate informative canonical definitions. Nevertheless, the evaluation on GSCplus with the Human Phenotype Ontology demonstrates that the approach can transfer beyond the food domain and FoodOn.

ReciLinkD focuses on 1{,}014 food material concepts extracted from selected FoodOn branches rather than the complete FoodOn ontology. This restriction provides a focused candidate space for recipe food entity linking, but concepts outside this space cannot occur as gold entities, and the benchmark should therefore not be interpreted as covering the complete FoodOn ontology, as non-food-related concepts are not covered. A restricted candidate space may also reduce entity linking difficulty by removing unrelated concepts. However, evaluation on GSCplus shows that the approach transfers to ontologies with larger candidate spaces, reaching 17{,}655 Ontology concepts. 

The main SCODEL experiments use manually constructed canonical definitions as training supervision. While this introduces an initial annotation requirement, canonical definitions can also be generated automatically using LLMs, with our experiments showing that LLM-generated definitions provide an effective alternative to manual supervision. 

\bibliography{custom}
\appendix

\section{Extracting Food Related Concepts from FoodOn Ontology}
\label{appendix:foodon_extract}
Algorithm~\ref{alg:foodon_extract} describes the procedure for extracting relevant food concepts from the FoodOn ontology. We select four root categories: \textit{fruit} (\texttt{PO\_0009001}), \textit{seed} (\texttt{PO\_0009010}), \textit{food (raw)} (\texttt{FOODON\_03311126}) and \textit{cultural food} (\texttt{FOODON\_00001796})). These roots restrict our selected concepts to FoodOn branches corresponding to food materials relevant to recipe entity linking. For each root, we retrieve all descendant concepts and identify leaf nodes within the resulting subtree. To capture both fine-grained and more general concepts, we include each leaf node together with its immediate parent class. We use a leaf-plus-parent extraction strategy to balance concept specificity and coverage. Leaf nodes provide the fine-grained food concepts required for precise entity linking, while their immediate parents retain broader concepts that may also correspond directly to food mentions in recipes. Finally, we normalise and deduplicate the resulting concept set to produce the candidate ontology set used for annotation. Table~\ref{tab:food_stats} summarises key statistics of the annotated ReciLinkD dataset.

\begin{algorithm}[ht!]
\caption{Steps for Extracting Food Concepts from FoodOn Ontology}
\label{alg:foodon_extract}
\begin{algorithmic}[1]
\REQUIRE FoodOn ontology $\mathcal{O}$
\ENSURE Candidate concept set $\mathcal{C}$

\STATE $\mathcal{R} \leftarrow$ 
\{\texttt{PO\_0009001}, \texttt{PO\_0009010}, 
\texttt{FOODON\_03311126}, \texttt{FOODON\_00001796}\}

\STATE $\mathcal{C} \leftarrow \emptyset$

\FOR{each parent $r \in \mathcal{R}$}
    \STATE $\mathcal{T} \leftarrow \text{Descendants}(r) \cup \{r\}$
    \STATE $\mathcal{L} \leftarrow \{c \in \mathcal{T} \mid \text{Children}_{\mathcal{T}}(c)=\emptyset\}$ \COMMENT{leaf nodes}
    \FOR{each $\ell \in \mathcal{L}$}
        \STATE $\mathcal{C} \leftarrow \mathcal{C} \cup \{\ell\}$
        \STATE $\mathcal{C} \leftarrow \mathcal{C} \cup \text{Parents}(\ell)$
    \ENDFOR
\ENDFOR

\STATE Normalise and deduplicate $\mathcal{C}$
\RETURN $\mathcal{C}$
\end{algorithmic}
\end{algorithm}

\begin{table}[ht!]
\centering
\small
\setlength{\tabcolsep}{5pt}
\begin{tabular}{|l|c|}
\hline
\textbf{Metric} & \textbf{Value} \\
\hline
Unique manually selected food mentions & 4{,}500 \\
Unique manually linked food mentions & 3{,}979 \\
Unique unlinked food mentions & 521 \\
Unique FoodOn food concepts & 979 \\
Rows with synonyms & 2,023 \\
Rows with superclasses & 2,970 \\
Rows with subclasses & 1,971 \\
Rows with direct parents & 1,649 \\
Rows with direct children & 1,151 \\
Avg. synonyms per row & 0.82 \\
Avg. parent concepts per row & 1.37 \\
\hline
\end{tabular}
\caption{Descriptive statistics for ReciLinkD.}
\label{tab:food_stats}
\end{table}

\section{Supplementary Ablation Studies}
\label{appendix:further_ablation_study}
\subsection{Choosing Canonical Definition Source}
Table~\ref{tab:definition_source_ablation} compares different sources for constructing the shared canonical training definitions. Human-annotated definitions achieve the best overall performance, indicating that carefully aligned canonical definitions provide the strongest supervision for SCODEL. Nevertheless, canonical definitions generated by both proprietary and open-weight LLMs consistently outperform the ReFinED baseline, suggesting that automatically generated shared definitions offer an effective and scalable alternative when manual annotation is impractical for large ontologies.

\begin{table}[ht!]
\centering
\small
\setlength{\tabcolsep}{8pt}
\begin{tabular}{|l|c|c|}
\hline
\textbf{Training Definition Source} &
\textbf{Hit@1} &
\textbf{MRR} \\
\hline
ReFinED Baseline & 81.82 & 85.42 \\
\hline
OpenAI GPT-5.1-2025-11-13 & \underline{84.91} & \underline{89.06} \\
Claude Sonnet 5 & 84.05 & 88.28 \\
Qwen3-235B & 83.36 & 88.07 \\
Human Annotated & \textbf{85.08} & \textbf{89.11} \\
\hline
\end{tabular}
\caption{Effect of different sources used for constructing the canonical definitions. All experiments use Mistral-7B as the shared definition learner and generator and ReFinED as the downstream entity linking model. Best results are bold and second-best results are underlined.}
\label{tab:definition_source_ablation}
\end{table}

\subsection{Semantic Alignment Analysis}
\label{app:semantic_alignment_analysis}
We examine whether shared canonical definition learning directly improves semantic alignment between recipe mentions and ontology concepts by evaluating definition-based concept retrieval before and after SCODEL training. In both settings, definitions are generated independently for the contextual mention and ontology metadata, embedded using the same \texttt{all-mpnet-base-v2} model, and ranked using cosine similarity over the same 979 candidate concepts. The baseline uses the original Mistral-7B in a zero-shot setting for definition generation as in LLM4EL \cite{xu2026llm4el}, whereas SCODEL uses Mistral-7B after shared canonical definition learning.

We additionally measure \textit{Gold Sim.}, the cosine similarity between the food mention definition and its gold ontology concept definition and \textit{Global Margin}, the difference between the gold similarity and the mean similarity to non-matching ontology concepts.

\begin{table}[ht!]
\centering
\small
\setlength{\tabcolsep}{4pt}
\begin{tabular}{|l|c|c|}
\hline
\textbf{Metric} &
\textbf{Mistral-7B} &
\textbf{SCODEL} \\
\hline
Hit@1 & 27.62 & \textbf{46.31} \\
\hline
Hit@5 & 51.63 & \textbf{70.84} \\
\hline
Hit@10 & 64.67 & \textbf{78.73} \\
\hline
MRR & 39.92 & \textbf{57.72} \\
\hline
Gold Sim. & 0.6771 & \textbf{0.8312} \\
\hline
% Hard Neg. Sim. & 0.7430 & 0.8690 \\
% \hline
Global Margin & 0.3148 & \textbf{0.3595} \\
\hline
% Hard Margin & $-0.0659$ & \textbf{$-0.0378$} \\
% \hline
\end{tabular}
\caption{Semantic alignment before and after SCODEL shared canonical definition learning. Mistral-7B denotes zero-shot definition generation using the original model as in LLM4EL~\cite{xu2026llm4el}. Best results are bold.}
\label{tab:semantic_alignment}
\end{table}

SCODEL improves retrieval across all ranking metrics, increasing Hit@1 from 27.62\% to 46.31\% and MRR from 39.92\% to 57.72\%. Gold similarity increases from 0.6771 to 0.8312, while the global margin increases from 0.3148 to 0.3595. These results provide direct evidence that shared canonical definition learning improves semantic alignment between recipe mention definitions and their corresponding ontology concept definitions.

\subsection{Choosing Definition Learning Backbone Size}
Model size influences the quality of the learned canonical definitions, as shown in Table~\ref{tab:model_size}. Performance generally improves with increasing model size, although the 8B model slightly underperforms the 3B variant. The 70B model achieves the best overall results, suggesting that larger language models learn more effective shared canonical definitions for downstream entity linking.

\begin{table}[ht!]
\centering
\small
\setlength{\tabcolsep}{8pt}
\begin{tabular}{|l|c|c|c|}
\hline
\textbf{Model Family} &
\textbf{Parameters} &
\textbf{Hit@1} &
\textbf{MRR} \\
\hline

LLaMA-3 & 1B  & 81.82 & 85.60 \\
\hline
LLaMA-3 & 3B  & \underline{83.53} & \underline{87.12} \\
\hline
LLaMA-3 & 8B  & 82.33 & 86.43 \\
\hline
LLaMA-3 & 70B & \textbf{84.91} & \textbf{88.72} \\
\hline

\end{tabular}
\caption{Effect of model size on SCODEL shared definition learning and generation. ReFinED is used as the downstream entity linking model. Best results are bold and second-best results are underlined.}
\label{tab:model_size}
\end{table}

\subsection{LLMAEL Across Entity Linking Models}
\label{app:llmael_other_el_models}
We assess LLMAEL's effectiveness across different entity linking architectures by applying it to three downstream entity linking models: ReFinED, GSM-EL, and XMR. Mistral was retained as the language model across all experiments to ensure that differences in performance primarily reflect the choice of entity linking model.

Table~\ref{tab:llmael_el_models} shows that LLMAEL performs best when ReFinED is used as the downstream entity linking model. The ReFinED configuration obtains a Hit@1 of 85.25\%, Hit@5 of 94.51\%, and MRR of 89.61\%, outperforming the corresponding GSM-EL and XMR configurations.

\begin{table}[ht!]
\centering
\small
\setlength{\tabcolsep}{8pt}
\begin{tabular}{|l|c|c|c|}
\hline
\textbf{Entity Linking Model} &
\textbf{Hit@1} &
\textbf{Hit@5} &
\textbf{MRR} \\
\hline

ReFinED & \textbf{85.25} & \textbf{94.51} & \textbf{89.61} \\
\hline
GSM-EL & 81.81 & 93.65 & 87.06 \\
\hline
BLINK & 85.22 & 93.82 & 89.60 \\
\hline
\end{tabular}
\caption{Performance of LLMAEL across different downstream entity linking models. Mistral-7B is used as the language model across all configurations. Best results are in bold.}
\label{tab:llmael_el_models}
\end{table}

\section{ReciLinkD Inter-Annotator Agreement}
\label{appendix:annotation_agreement}

To assess the reliability of the ReciLinkD annotations, 300 randomly selected recipe food mentions were independently annotated by all four annotators. Agreement was evaluated across all six annotator pairs using pairwise F1-score and Cohen's $\kappa$. Table~\ref{tab:annotation_agreement} reports the results for each annotator pair.

\begin{table}[ht!]
\centering
\small
\setlength{\tabcolsep}{8pt}
\begin{tabular}{|l|c|c|}
\hline
\textbf{Annotator Pair} &
\textbf{F1 (\%)} &
\textbf{Cohen's $\boldsymbol{\kappa}$} \\
\hline
A--B & 85.49 & 0.7178 \\
\hline
A--C & 90.56 & 0.7254 \\
\hline
A--D & 91.31 & 0.7404 \\
\hline
B--C & 89.53 & 0.7219 \\
\hline
B--D & 88.68 & 0.7203 \\
\hline
C--D & 92.43 & 0.7258 \\
\hline
\textbf{Mean} & \textbf{89.67} & \textbf{0.7253} \\
\hline
\end{tabular}
\caption{Pairwise inter-annotator agreement for the 300 ReciLinkD food mentions independently annotated by all four annotators.}
\label{tab:annotation_agreement}
\end{table}

The pairwise F1-scores range from 85.49\% to 92.43\%, with a mean of 89.67\%. Cohen's $\kappa$ ranges from 0.7178 to 0.7404, with a mean of 0.7253. These results indicate consistently strong agreement across annotator pairs.

\section{Large-Scale Entity Linking to ReciFine}
\label{appendix:recifine_linking}
We extend ReciLinkD to a substantially larger set of food mentions from ReciFine~\cite{nuhuibrahim2024recifine}, extracted using the methods in \citet{ibrahim2026knowledge}, to evaluate SCODEL beyond the benchmark and build a larger manually verified resource for ontology grounding. The process selects additional ReciFine food mentions, links them to FoodOn using SCODEL $\times$ ReFinED, manually verifies the predictions, and combines the retained mappings with ReciLinkD.

\subsection{Selection and Linking}
ReciFine contains food mentions extracted from instructions of approximately 2.2M recipes. We rank these food mentions by the frequency of their appearance in recipes and select the 20{,}000 most frequent food mentions not already covered by ReciLinkD. Frequency-based selection prioritises mappings with greater potential corpus coverage, since a single verified mapping can ground multiple occurrences of the same food mention across recipes in ReciFine.
 
\subsection{Manual Verification}
All 20{,}000 predicted food mention and ontology concept pairs are manually reviewed. Each prediction is assigned one of three outcomes: \textit{Correct and retained}, where the expression is a valid food mention and the predicted FoodOn concept represents the intended entity; \textit{Incorrect FoodOn prediction}, where the expression is a valid or invalid food mention and is linked to an incorrect FoodOn concept; or \textit{Correct unsuitable food mention}, where the predicted concept is contextually appropriate, but the extracted expression itself is unsuitable as a standalone food name.

The final category separates entity-linking errors from food-mention extraction incompatibilities. For example, in \textit{mix the flour, butter, and sugar until a crumbly mixture forms, then spread it over the fruit}, the referent of \textit{it} can be identified from the recipe context as the \textit{dough}, although the extracted mention \textit{it} does not explicitly name the food entity. The entity-linking prediction may therefore be correct, but retaining \textit{it} as a lexical alias for the corresponding FoodOn concept would introduce noise and inconsistencies when the resource is subsequently propagated across ReciFine.

Table~\ref{tab:recifine_manual_validation} reports the manual evaluation results.

\begin{table}[ht!]
\centering
\small
\setlength{\tabcolsep}{8pt}
\begin{tabular}{|l|r|r|}
\hline
\textbf{Manual Evaluation Outcome} &
\textbf{Count} &
\textbf{Perc.} \\
\hline
Correct and retained & 11,580 & 57.90\% \\
\hline
Correct unsuitable food mention & 7{,}226 & 36.13\% \\
\hline
Incorrect FoodOn prediction & 2{,}194 & 10.97\% \\
\hline
\end{tabular}
\caption{Manual evaluation of SCODEL $\times$ ReFinED predictions for 20{,}000 additional ReciFine food mentions.}
\label{tab:recifine_manual_validation}
\end{table}

Among valid food mentions, SCODEL $\times$ ReFinED correctly links 18{,}806 to the appropriate FoodOn concept. After excluding expressions unsuitable for subsequent lexical grounding, 11{,}580 additional manually verified food mention and ontology concept mappings are retained.

\subsection{Expanded Linking Resource}
The new retained ReciFine mappings are combined with the original ReciLinkD annotations to form an expanded manually verified food mention and ontology concept resource. Table~\ref{tab:expanded_linking_resource} reports its resulting coverage.

\begin{table}[ht!]
\centering
\small
\setlength{\tabcolsep}{8pt}
\begin{tabular}{|l|r|}
\hline
\textbf{Source} &
\textbf{Mention--Concept Pairs} \\
\hline
ReciLinkD & 3,979 \\
\hline
ReciLinkD Expansion & 11{,}580 \\
\hline
\textbf{Combined resource} & \textbf{15{,}559} \\
\hline
\end{tabular}
\caption{Composition of the expanded manually verified food mention and ontology concept linking resource.}
\label{tab:expanded_linking_resource}
\end{table}

\section{Ontology-Grounded Recipe Knowledge Graph}
\label{appendix:kg_details}
The expanded linking resource provides verified correspondences between ReciFine food mention surface forms and FoodOn concepts. We use these mappings to ground food mentions across the wider ReciFine corpus and integrate the resulting recipe entities with the FoodOn hierarchy.

\subsection{Vocabulary Expansion and ReciFine Grounding}
To propagate the manually verified food mention and FoodOn concept mappings efficiently, we construct a concept-preserving lexical vocabulary for each represented FoodOn concept. The vocabulary combines manually verified food mentions with the corresponding FoodOn concept label, synonyms, and their conservative variants such as singular and plural forms.

For example, a verified mapping from \textit{chickpeas} to a concept with the preferred label \textit{chickpea} and synonym \textit{garbanzo bean} allows these equivalent lexical forms, together with conservative plural variants such as \textit{garbanzo beans}, to ground corresponding food mentions in ReciFine to the same FoodOn concept.

The expansion remains concept-preserving: labels and synonyms are used only as aliases for the FoodOn concept to which they belong. Parent and child labels are not treated as lexical aliases, since ontological relatedness does not imply lexical equivalence. Where the same normalised form is associated with multiple FoodOn concepts, the ambiguity is retained rather than randomly assigning the expression to one concept.

Importantly, this propagation stage does not introduce new SCODEL predictions. Instead, it expands the coverage of manually verified mappings using lexical forms associated with their established FoodOn concepts, allowing additional ReciFine food mentions to be grounded correctly despite differences in surface form or spelling. Table~\ref{tab:alias_expansion} shows the resulting vocabulary expansion.

\begin{table}[ht!]
\centering
\small
\setlength{\tabcolsep}{8pt}
\begin{tabular}{|l|r|}
\hline
\textbf{Vocabulary Expansion} & \textbf{Count} \\
\hline
Original verified food mention names & 15{,}559 \\
\hline
FoodOn concept labels & 979 \\
\hline
Names from FoodOn concept synonyms & 253 \\
\hline
Singular/plural name variants & 16{,}738 \\
\hline
\textbf{Final expanded vocabulary} & \textbf{33{,}529} \\
\hline
\end{tabular}
\caption{Expansion of the verified food-name vocabulary used to ground additional ReciFine mentions.}
\label{tab:alias_expansion}
\end{table}

We use the resulting vocabulary to ground the food mentions previously extracted from the ReciFine recipes. We match each extracted food mention against the vocabulary and, when we find a valid match, assign the corresponding FoodOn concept. This propagates the manually verified mappings across ReciFine, producing ontology-grounded food mentions that we then use to construct the recipe knowledge graph. Table~\ref{tab:recifine_grounding} summarises the resulting corpus coverage.

\begin{table}[ht!]
\centering
\scriptsize
\setlength{\tabcolsep}{8pt}
\begin{tabular}{|l|r|}
\hline
\textbf{Statistic} & \textbf{Value} \\
\hline
ReciFine recipes considered & 2,206,253 \\
\hline
% Unique extracted food mentions & 307,917 \\
% \hline
Unique food mentions matched to vocabulary & 19,504 \\
\hline
% Food mentions remaining unmatched & 288,413 \\
% \hline
Recipes grounded to food mentions & 2,062,103 \\
\hline
\end{tabular}
\caption{Coverage obtained by propagating the expanded verified food mention and ontology concept vocabulary across ReciFine.}
\label{tab:recifine_grounding}
\end{table}

\subsection{FoodOn Hierarchy Integration}
Lexical grounding identifies the FoodOn concept corresponding to a recipe food mention, while the FoodOn hierarchy lets these concepts be interpreted through broader semantic categories. For each directly grounded concept, we therefore retain its FoodOn identifier and taxonomic relationships.

If a recipe contains a mention linked to concept \(c\), and \(c\) is a descendant of concept \(a\), the recipe can also be interpreted and retrieved as containing an instance of \(a\). This distinction between \textit{direct grounding} and \textit{inherited category membership} is maintained throughout the graph: the former records the concept explicitly identified from a recipe mention, whereas the latter is inferred through FoodOn's ontology structure.

The resulting graph contains three principal node types: recipe nodes, food mention nodes, and FoodOn concept nodes. Its principal structure is

\[
r
\xrightarrow{\texttt{contains}}
m
\xrightarrow{\texttt{linked\_to}}
c
\xrightarrow{\texttt{is\_a}^{*}}
a,
\]

where \(r\) denotes a recipe, \(m\) a grounded food mention, \(c\) its directly linked FoodOn concept, and \(a\) an inherited higher-level FoodOn concept.

\subsection{Knowledge Graph Statistics}
Table~\ref{tab:kg_statistics} reports the resulting graph statistics. We distinguish directly grounded FoodOn concepts from ontology concepts introduced through hierarchy expansion to avoid conflating concepts observed in recipes with those included only as semantic context.

\begin{table}[ht!]
\centering
\scriptsize
\setlength{\tabcolsep}{8pt}
\begin{tabular}{|l|r|}
\hline
\textbf{Knowledge Graph Statistic} & \textbf{Value} \\
\hline
ReciFine recipes considered & 2,206,253 \\
\hline
% Recipe nodes & 2,206,253 \\
% \hline
Grounded recipe nodes & 2,062,103 \\
\hline
Grounded food mention nodes & 14,177,860 \\
\hline
Directly grounded FoodOn concepts & 979 \\
\hline
Inherited FoodOn concepts & 2,209 \\
\hline
Recipe $\rightarrow$ mention edges & 14,177,860 \\
\hline
Food mention $\rightarrow$ concept edges & 14,177,860 \\
\hline
Direct ontology hierarchy edges & 4,190 \\
\hline
Derived ontology closure edges & 17,555 \\
\hline
\textbf{Total grounded principal graph nodes} & \textbf{16,243,151} \\
\hline
\textbf{Total grounded principal graph edges} & \textbf{28,359,910} \\
\hline
\end{tabular}
\caption{Statistics of the expanded ontology-grounded ReciFine knowledge graph. Of the 2,206,253 ReciFine recipes considered, 2,062,103 contain at least one grounded food mention and are represented as recipe nodes. Derived transitive ontology closure edges are reported separately and excluded from the principal graph-edge total.}
\label{tab:kg_statistics}
\end{table}

\section{Ontology-Aware Retrieval and Corpus Analysis}
\label{appendix:kg_analysis}
We evaluate the practical utility of the ontology-grounded representation through recipe retrieval and corpus-level semantic analysis. The former examines whether FoodOn hierarchy information improves retrieval beyond lexical and semantic search. Retrieval relevance is assessed using an LLM-as-a-judge, with two human annotators independently evaluating a sample to measure inter-annotator and human--LLM agreement. The latter examines food-category prevalence and co-occurrence across ReciFine using previously verified food mention mappings to gold FoodOn concepts, allowing corpus-level analysis without additional manual annotation.

\subsection{Ontology-Aware Recipe Retrieval}
\label{appendix:ontology_aware_recipe_retrieval}
Ontology-aware retrieval identifies the corresponding FoodOn concept and retrieves recipes containing either that concept or any of its descendants. Thus, a query for \textit{fruit}, for example, can retrieve recipes grounded to specific descendant concepts such as \textit{apple}, \textit{banana}, \textit{strawberry}, and \textit{blueberry} even when the lexical form \textit{fruit} does not occur in the recipe.

The retrieval search space consists of 2{,}000 recipes selected from the approximately 2.2M recipes in ReciFine using proportionate stratified random sampling. To construct a corpus-representative evaluation collection, we stratify recipes by the number of food mentions grounded to verified FoodOn concepts and by recipe instruction length. We sample approximately proportional numbers of recipes from each stratum, using a fixed random seed for reproducibility. This ensures that the evaluation includes recipes with varying levels of ontology coverage (including recipes with no food mentions grounded to FoodOn concepts) and textual complexity. The same fixed 2,000-recipe collection is used to evaluate lexical, semantic, and ontology-aware retrieval.

For lexical retrieval, we normalise the query and then retrieve recipes that contain a food mention that directly matches the query. For example, a query for \textit{fruit} retrieves recipes that explicitly contain the mention \textit{fruit}, but not recipes that contain only \textit{apple} or \textit{banana}.

For semantic retrieval, we encode the query and extracted food mentions independently using the pretrained \texttt{all-mpnet-base-v2} sentence-transformer model and compute their cosine similarity. Recipes containing food mentions whose cosine similarity to the query is at least a fixed threshold of \textit{0.70} are retrieved. This allows semantic retrieval to identify related food mentions without requiring an exact lexical match or using ontology information.

For ontology-aware retrieval, we map the query to its corresponding FoodOn concept and retrieve recipes grounded to either that concept or any of its descendants. Thus, a query for \textit{fruit} can retrieve recipes grounded to specific descendant concepts such as \textit{apple}, \textit{banana}, \textit{strawberry}, and \textit{blueberry}, even when the word \textit{fruit} does not occur in the recipe. Unlike semantic retrieval, this approach determines category membership from explicit relationships in the FoodOn hierarchy rather than embedding similarity.

\begin{figure}[ht!]
\centering
\includegraphics[width=0.35\textwidth]{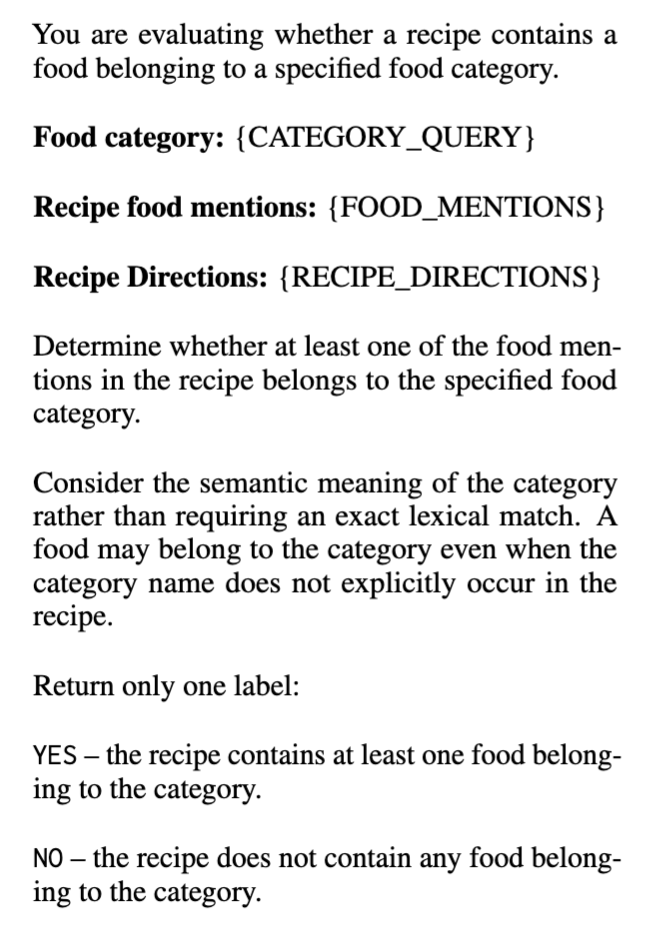}
\caption{Prompt template for determining whether the recipe contains food mentions belonging to the queried category.}
\label{fig:prompt_auto_category_for_retrieval}
\end{figure}

We evaluate the three approaches independently across the same 15 hand-selected higher-level food-category queries. To enable evaluation at scale, retrieved recipes are automatically assessed using a GPT-5.6-Sol model. Given a food-category query and the food mentions extracted from a recipe, the model determines whether the recipe contains at least one food mention belonging to the queried category. This produces a binary relevance judgement for each evaluated query and recipe pair. Figure~\ref{fig:prompt_auto_category_for_retrieval} illustrates the prompt used to determine whether the recipe contains food mentions belonging to the queried category.

To assess the reliability of these automatic judgements, we conduct a separate human agreement study on 200 randomly sampled query and recipe pairs. Two human annotators independently assign binary relevance labels to each pair using the same criterion as the automatic annotator. We measure inter-annotator agreement between the two human annotators and agreement between the GPT-5.6-Sol annotations and each human annotator. The resulting agreement scores are reported in Table~\ref{tab:retrieval_annotation_agreement}. Human annotators show substantial agreement ($\kappa=0.76$), while GPT-5.6-Sol also achieves substantial agreement with both annotators ($\kappa=0.64$ and $0.61$). 

For each retrieval approach and query, we report precision, recall, and F1-score based on the automatic relevance judgements in Table~\ref{tab:ontology_retrieval}. A recipe is considered relevant when its food entities satisfy the semantic category represented by the query.

\begin{table}[ht!]
\centering
\tiny
\setlength{\tabcolsep}{8pt}
\begin{tabular}{|l|c|c|}
\hline
\textbf{Annotator Pair} &
\textbf{Agreement} &
\textbf{Cohen's $\kappa$} \\
\hline
Human 1 -- Human 2 & 88.95\% & 0.76 \\
\hline
Human 1 -- GPT-5.6-Sol & 82.00\% & 0.64 \\
\hline
Human 2 -- GPT-5.6-Sol & 80.50\% & 0.61 \\
\hline
\end{tabular}
\caption{Agreement on 200 randomly sampled query and recipe relevance judgements. Agreement denotes the percentage of identical binary relevance labels, while Cohen's kappa ($\kappa$) accounts for chance agreement.}
\label{tab:retrieval_annotation_agreement}
\end{table}

\begin{table}[ht!]
\centering
\tiny
\setlength{\tabcolsep}{2pt}
\begin{tabular}{|l|ccc|ccc|ccc|}
\hline
\multirow{2}{*}{\textbf{Query}} &
\multicolumn{3}{c|}{\textbf{Lexical}} &
\multicolumn{3}{c|}{\textbf{Semantic}} &
\multicolumn{3}{c|}{\textbf{Ontology}} \\
\cline{2-10}
&
\textbf{P} & \textbf{R} & \textbf{F1} &
\textbf{P} & \textbf{R} & \textbf{F1} &
\textbf{P} & \textbf{R} & \textbf{F1} \\
\hline
\href{https://ontobee.org/ontology/FOODON?iri=http://purl.obolibrary.org/obo/PO_0009001}{Fruit} & 86.67 & 4.34 & 8.26 & 91.26 & 10.46 & 18.76 & 71.77 & 76.08 & \textbf{73.87} \\
\hline
\href{https://ontobee.org/ontology/FOODON?iri=http://purl.obolibrary.org/obo/FOODON_03302008}{Vegetable} & 85.11 & 3.45 & 6.64 & 96.19 & 8.72 & 15.99 & 82.06 & 18.57 & \textbf{30.28} \\
\hline
\href{https://ontobee.org/ontology/FOODON?iri=http://purl.obolibrary.org/obo/FOODON_03310646}{Legume} & 0.00 & 0.00 & 0.00 & 0.00 & 0.00 & 0.00 & 60.37 & 52.11 & \textbf{55.93} \\
\hline
\href{https://ontobee.org/ontology/FOODON?iri=http://purl.obolibrary.org/obo/FOODON_03306429}{Meat} & 80.25 & 20.93 & 33.21 & 85.64 & 26.89 & 40.93 & 94.76 & 34.94 & \textbf{51.06} \\
\hline
\href{https://ontobee.org/ontology/FOODON?iri=http://purl.obolibrary.org/obo/FOODON_02021659}{Pork} & 91.49 & 21.83 & \textbf{35.25} & 91.49 & 21.83 & \textbf{35.25} & 88.24 & 15.23 & 25.97 \\
\hline
\href{https://ontobee.org/ontology/FOODON?iri=http://purl.obolibrary.org/obo/FOODON_03311865}{Chicken} & 97.64 & 71.13 & \textbf{82.31} & 97.70 & 58.42 & 73.12 & 96.84 & 63.23 & 76.51 \\
\hline
\href{https://ontobee.org/ontology/FOODON?iri=http://purl.obolibrary.org/obo/FOODON_03310116}{Seafood} & 100.00 & 1.38 & 2.72 & 95.16 & 40.69 & \textbf{57.00} & 97.06 & 22.76 & 36.87 \\
\hline
\href{https://ontobee.org/ontology/FOODON?iri=http://purl.obolibrary.org/obo/FOODON_03301088}{Crustacean} & 0.00 & 0.00 & 0.00 & 0.00 & 0.00 & 0.00 & 94.12 & 49.23 & \textbf{64.65} \\
\hline
\href{https://ontobee.org/ontology/FOODON?iri=http://purl.obolibrary.org/obo/FOODON_00003520}{Pepper} & 40.70 & 51.98 & 45.65 & 43.83 & 60.95 & 50.99 & 80.34 & 37.73 & \textbf{51.35} \\
\hline
\href{https://ontobee.org/ontology/FOODON?iri=http://purl.obolibrary.org/obo/FOODON_00003946}{Winter squash} & 0.00 & 0.00 & 0.00 & 29.41 & 16.13 & 20.83 & 95.00 & 61.29 & \textbf{74.51} \\
\hline
\href{https://ontobee.org/ontology/FOODON?iri=http://purl.obolibrary.org/obo/PO_0030110}{Pome fruit} & 0.00 & 0.00 & 0.00 & 0.00 & 0.00 & 0.00 & 98.28 & 61.96 & \textbf{76.00} \\
\hline
\href{https://ontobee.org/ontology/FOODON?iri=http://purl.obolibrary.org/obo/PO_0030103}{Drupe fruit} & 0.00 & 0.00 & 0.00 & 0.00 & 0.00 & 0.00 & 41.38 & 17.96 & \textbf{25.04} \\
\hline
\href{https://ontobee.org/ontology/FOODON?iri=http://purl.obolibrary.org/obo/FOODON_00003592}{Aggregate fruit} & 0.00 & 0.00 & 0.00 & 0.00 & 0.00 & 0.00 & 90.00 & 54.55 & \textbf{67.92} \\
\hline
\href{https://ontobee.org/ontology/FOODON?iri=http://purl.obolibrary.org/obo/PO_0030104}{Cereal grain} & 0.00 & 0.00 & 0.00 & 84.21 & 1.58 & 3.09 & 92.42 & 6.01 & \textbf{11.29} \\
\hline
\href{https://ontobee.org/ontology/FOODON?iri=http://purl.obolibrary.org/obo/PO_0030102}{Nut} & 100.00 & 2.13 & 4.17 & 45.54 & 19.57 & \textbf{27.38} & 100.00 & 1.70 & 3.35 \\

\hline
\textbf{Average} &
45.46 & 11.81 & 14.55 &
50.70 & 17.68 & 22.89 &
\textbf{85.51} & \textbf{38.22} & \textbf{48.31} \\
\hline
\end{tabular}
\caption{Comparison of lexical, semantic, and ontology-aware recipe retrieval across 15 higher-level food-category queries. P, R, and F1 denote precision, recall, and F1-score (\%), respectively, over the 2{,}000-recipe ReciFine evaluation sample. The Average row reports averaged performance across the 15 queries. Best results are in bold.}
\label{tab:ontology_retrieval}
\end{table}

\subsection{Food-Category Prevalence}
\label{appendix:food_category_prevalence}
The Ontology hierarchy represented by the knowledge graph supports corpus-level analysis that would be difficult to perform reliably using individual food names alone. We use the FoodOn hierarchy to group directly grounded food concepts under broader food categories and then calculate how many recipes contain each category. A recipe is counted for a category when it contains at least one grounded FoodOn concept corresponding to that category or one of its descendants.

For example, recipes grounded to specific fruits such as apple, banana, or strawberry can all be counted under the broader \textit{fruit} category. The same process lets us determine the prevalence of broader categories such as vegetables, fish, dairy products, and animal- and plant-derived foods across ReciFine without manually specifying every food in each category, as shown in Table~\ref{tab:category_prevalence}.

\begin{table}[ht!]
\centering
\tiny
\setlength{\tabcolsep}{7pt}
\begin{tabular}{|l|r|r|}
\hline
\textbf{FoodOn Category} &
\textbf{Recipes} &
\textbf{Prevalence} \\
\hline
\href{https://ontobee.org/ontology/FOODON?iri=http://purl.obolibrary.org/obo/FOODON_00001015}{Plant food product} & 1{,}744{,}295 & 84.59\% \\
\hline
\href{https://ontobee.org/ontology/FOODON?iri=http://purl.obolibrary.org/obo/FOODON_00004242}{Animal food product} & 1{,}510{,}986 & 73.27\% \\
\hline
\href{https://ontobee.org/ontology/FOODON?iri=http://purl.obolibrary.org/obo/FOODON_00001261}{Vegetable food product} & 1{,}297{,}374 & 62.92\% \\
\hline
\href{https://ontobee.org/ontology/FOODON?iri=http://purl.obolibrary.org/obo/FOODON_00001256}{Dairy food product} & 1{,}046{,}656 & 50.76\% \\
\hline
\href{https://ontobee.org/ontology/FOODON?iri=http://purl.obolibrary.org/obo/PO_0009001}{Fruit} & 978{,}283 & 47.44\% \\
\hline
\href{https://ontobee.org/ontology/FOODON?iri=http://purl.obolibrary.org/obo/FOODON_00001133}{Condiment food product} & 939{,}353 & 45.55\% \\
\hline
\href{https://ontobee.org/ontology/FOODON?iri=http://purl.obolibrary.org/obo/FOODON_00001242}{Spice or herb} & 742{,}511 & 36.01\% \\
\hline
\href{https://ontobee.org/ontology/FOODON?iri=http://purl.obolibrary.org/obo/FOODON_00002300}{Sweetener food product} & 690{,}735 & 33.50\% \\
\hline
\href{https://ontobee.org/ontology/FOODON?iri=http://purl.obolibrary.org/obo/FOODON_03460177}{Plant seed or nut food product} & 622{,}813 & 30.20\% \\
\hline
\href{https://ontobee.org/ontology/FOODON?iri=http://purl.obolibrary.org/obo/FOODON_00002664}{Lipid food product} & 575{,}016 & 27.88\% \\
\hline
\href{https://ontobee.org/ontology/FOODON?iri=http://purl.obolibrary.org/obo/FOODON_00001251}{Avian animal food product} & 551{,}230 & 26.73\% \\
\hline
\href{https://ontobee.org/ontology/FOODON?iri=http://purl.obolibrary.org/obo/FOODON_00001274}{Egg food product} & 508{,}606 & 24.66\% \\
\hline
\href{https://ontobee.org/ontology/FOODON?iri=http://purl.obolibrary.org/obo/FOODON_00001006}{Mammalian meat food product} & 503{,}119 & 24.40\% \\
\hline
\href{https://ontobee.org/ontology/FOODON?iri=http://purl.obolibrary.org/obo/FOODON_03311737}{Processed food product} & 479{,}192 & 23.24\% \\
\hline
\href{https://ontobee.org/ontology/FOODON?iri=http://purl.obolibrary.org/obo/FOODON_00001931}{Gravy or sauce} & 389{,}712 & 18.90\% \\
\hline
\href{https://ontobee.org/ontology/FOODON?iri=http://purl.obolibrary.org/obo/FOODON_00001093}{Cereal grain food product} & 280{,}506 & 13.60\% \\
\hline
\href{https://ontobee.org/ontology/FOODON?iri=http://purl.obolibrary.org/obo/FOODON_00001151}{Citrus fruit food product} & 264{,}798 & 12.84\% \\
\hline
\href{https://ontobee.org/ontology/FOODON?iri=http://purl.obolibrary.org/obo/FOODON_00003042}{Herb food product} & 257{,}504 & 12.49\% \\
\hline
\href{https://ontobee.org/ontology/FOODON?iri=http://purl.obolibrary.org/obo/FOODON_00001258}{Food (fermented)} & 257{,}377 & 12.48\% \\
\hline
\href{https://ontobee.org/ontology/FOODON?iri=http://purl.obolibrary.org/obo/FOODON_00001264}{Legume food product} & 228{,}828 & 11.10\% \\
\hline
\end{tabular}
\caption{Recipe-level prevalence of 20 selected higher-level FoodOn categories obtained through ontology hierarchy propagation on the 2{,}062{,}103 grounded recipes. Categories were selected for culinary interpretability, diversity, and to reduce redundancy between closely related ontology concepts.}
\label{tab:category_prevalence}
\end{table}

This analysis demonstrates how ontology-grounded entity linking enables individual food mentions to be analysed at broader semantic levels. The resulting category-level information can support further analyses of dietary composition and nutritional patterns and, when combined with external environmental data, the environmental characteristics of recipe ingredients.

\subsection{Food-Category Co-occurrence}
\label{appendix:food_category_cooccurence}
Ontology grounding also lets us examine how higher-level food categories co-occur across recipes. For each pair of categories, we count the number of recipes containing both categories and calculate the proportion of recipes containing the first category that also contain the second. For example, this shows how often recipes containing fish also contain vegetables, or how often plant- and animal-derived foods appear in the same recipe, as shown in Table~\ref{tab:category_cooccurrence}.

\begin{table}[ht!]
\centering
\tiny
\setlength{\tabcolsep}{6pt}
\begin{tabular}{|l|l|r|r|}
\hline
\textbf{Category A} &
\textbf{Category B} &
\textbf{Recipes} &
\textbf{Proportion} \\
\hline
\href{https://ontobee.org/ontology/FOODON?iri=http://purl.obolibrary.org/obo/FOODON_00001015}{Plant food product} &
\href{https://ontobee.org/ontology/FOODON?iri=http://purl.obolibrary.org/obo/FOODON_00004242}{Animal food product} &
1,303,624 & 74.74\% \\
\hline
\href{https://ontobee.org/ontology/FOODON?iri=http://purl.obolibrary.org/obo/FOODON_00004242}{Animal food product} &
\href{https://ontobee.org/ontology/FOODON?iri=http://purl.obolibrary.org/obo/FOODON_00001261}{Vegetable food product} &
969,055 & 64.13\% \\
\hline
\href{https://ontobee.org/ontology/FOODON?iri=http://purl.obolibrary.org/obo/FOODON_00001015}{Plant food product} &
\href{https://ontobee.org/ontology/FOODON?iri=http://purl.obolibrary.org/obo/FOODON_00001256}{Dairy food product} &
917,684 & 52.61\% \\
\hline
\href{https://ontobee.org/ontology/FOODON?iri=http://purl.obolibrary.org/obo/FOODON_00001261}{Vegetable food product} &
\href{https://ontobee.org/ontology/FOODON?iri=http://purl.obolibrary.org/obo/PO_0009001}{Fruit} &
891,116 & 68.69\% \\
\hline
\href{https://ontobee.org/ontology/FOODON?iri=http://purl.obolibrary.org/obo/FOODON_00001015}{Plant food product} &
\href{https://ontobee.org/ontology/FOODON?iri=http://purl.obolibrary.org/obo/FOODON_00001133}{Condiment food product} &
831,885 & 47.69\% \\
\hline
\href{https://ontobee.org/ontology/FOODON?iri=http://purl.obolibrary.org/obo/FOODON_00004242}{Animal food product} &
\href{https://ontobee.org/ontology/FOODON?iri=http://purl.obolibrary.org/obo/FOODON_00001133}{Condiment food product} &
746,545 & 49.41\% \\
\hline
\href{https://ontobee.org/ontology/FOODON?iri=http://purl.obolibrary.org/obo/FOODON_00004242}{Animal food product} &
\href{https://ontobee.org/ontology/FOODON?iri=http://purl.obolibrary.org/obo/PO_0009001}{Fruit} &
718,790 & 47.57\% \\
\hline
\href{https://ontobee.org/ontology/FOODON?iri=http://purl.obolibrary.org/obo/FOODON_00001133}{Condiment food product} &
\href{https://ontobee.org/ontology/FOODON?iri=http://purl.obolibrary.org/obo/FOODON_00002300}{Sweetener food product} &
690,735 & 73.53\% \\
\hline
\href{https://ontobee.org/ontology/FOODON?iri=http://purl.obolibrary.org/obo/FOODON_00001261}{Vegetable food product} &
\href{https://ontobee.org/ontology/FOODON?iri=http://purl.obolibrary.org/obo/FOODON_00001256}{Dairy food product} &
668,829 & 51.55\% \\
\hline
\href{https://ontobee.org/ontology/FOODON?iri=http://purl.obolibrary.org/obo/FOODON_00001261}{Vegetable food product} &
\href{https://ontobee.org/ontology/FOODON?iri=http://purl.obolibrary.org/obo/FOODON_00001242}{Spice or herb} &
616,146 & 47.49\% \\
\hline
\href{https://ontobee.org/ontology/FOODON?iri=http://purl.obolibrary.org/obo/FOODON_00001261}{Vegetable food product} &
\href{https://ontobee.org/ontology/FOODON?iri=http://purl.obolibrary.org/obo/FOODON_00001133}{Condiment food product} &
588,925 & 45.39\% \\
\hline
\href{https://ontobee.org/ontology/FOODON?iri=http://purl.obolibrary.org/obo/FOODON_00004242}{Animal food product} &
\href{https://ontobee.org/ontology/FOODON?iri=http://purl.obolibrary.org/obo/FOODON_00001242}{Spice or herb} &
579,969 & 38.38\% \\
\hline
\href{https://ontobee.org/ontology/FOODON?iri=http://purl.obolibrary.org/obo/FOODON_00004242}{Animal food product} &
\href{https://ontobee.org/ontology/FOODON?iri=http://purl.obolibrary.org/obo/FOODON_00002300}{Sweetener food product} &
552,911 & 36.59\% \\
\hline
\href{https://ontobee.org/ontology/FOODON?iri=http://purl.obolibrary.org/obo/FOODON_00001256}{Dairy food product} &
\href{https://ontobee.org/ontology/FOODON?iri=http://purl.obolibrary.org/obo/FOODON_00001133}{Condiment food product} &
520,590 & 49.74\% \\
\hline
\href{https://ontobee.org/ontology/FOODON?iri=http://purl.obolibrary.org/obo/FOODON_00001256}{Dairy food product} &
\href{https://ontobee.org/ontology/FOODON?iri=http://purl.obolibrary.org/obo/PO_0009001}{Fruit} &
508,850 & 48.62\% \\
\hline
\href{https://ontobee.org/ontology/FOODON?iri=http://purl.obolibrary.org/obo/PO_0009001}{Fruit} &
\href{https://ontobee.org/ontology/FOODON?iri=http://purl.obolibrary.org/obo/FOODON_00001133}{Condiment food product} &
502,324 & 51.35\% \\
\hline
\href{https://ontobee.org/ontology/FOODON?iri=http://purl.obolibrary.org/obo/PO_0009001}{Fruit} &
\href{https://ontobee.org/ontology/FOODON?iri=http://purl.obolibrary.org/obo/FOODON_00001242}{Spice or herb} &
460,548 & 47.08\% \\
\hline
\href{https://ontobee.org/ontology/FOODON?iri=http://purl.obolibrary.org/obo/FOODON_00001256}{Dairy food product} &
\href{https://ontobee.org/ontology/FOODON?iri=http://purl.obolibrary.org/obo/FOODON_00002300}{Sweetener food product} &
396,768 & 37.91\% \\
\hline
\href{https://ontobee.org/ontology/FOODON?iri=http://purl.obolibrary.org/obo/FOODON_00001256}{Dairy food product} &
\href{https://ontobee.org/ontology/FOODON?iri=http://purl.obolibrary.org/obo/FOODON_00001242}{Spice or herb} &
388,056 & 37.08\% \\
\hline
\href{https://ontobee.org/ontology/FOODON?iri=http://purl.obolibrary.org/obo/PO_0009001}{Fruit} &
\href{https://ontobee.org/ontology/FOODON?iri=http://purl.obolibrary.org/obo/FOODON_00002300}{Sweetener food product} &
375,448 & 38.38\% \\
\hline
\end{tabular}
\caption{Selected co-occurrence relationships between higher-level FoodOn categories. Proportion denotes the percentage of recipes containing Category A that also contain Category B.}
\label{tab:category_cooccurrence}
\end{table}

\section{Prompt for Shared Food Mention and Ontology Concept Pair Definition Generation}
\label{appendix:prompt_template_for_original_definition}
\begin{figure}[H]
\centering
\includegraphics[width=0.35\textwidth]{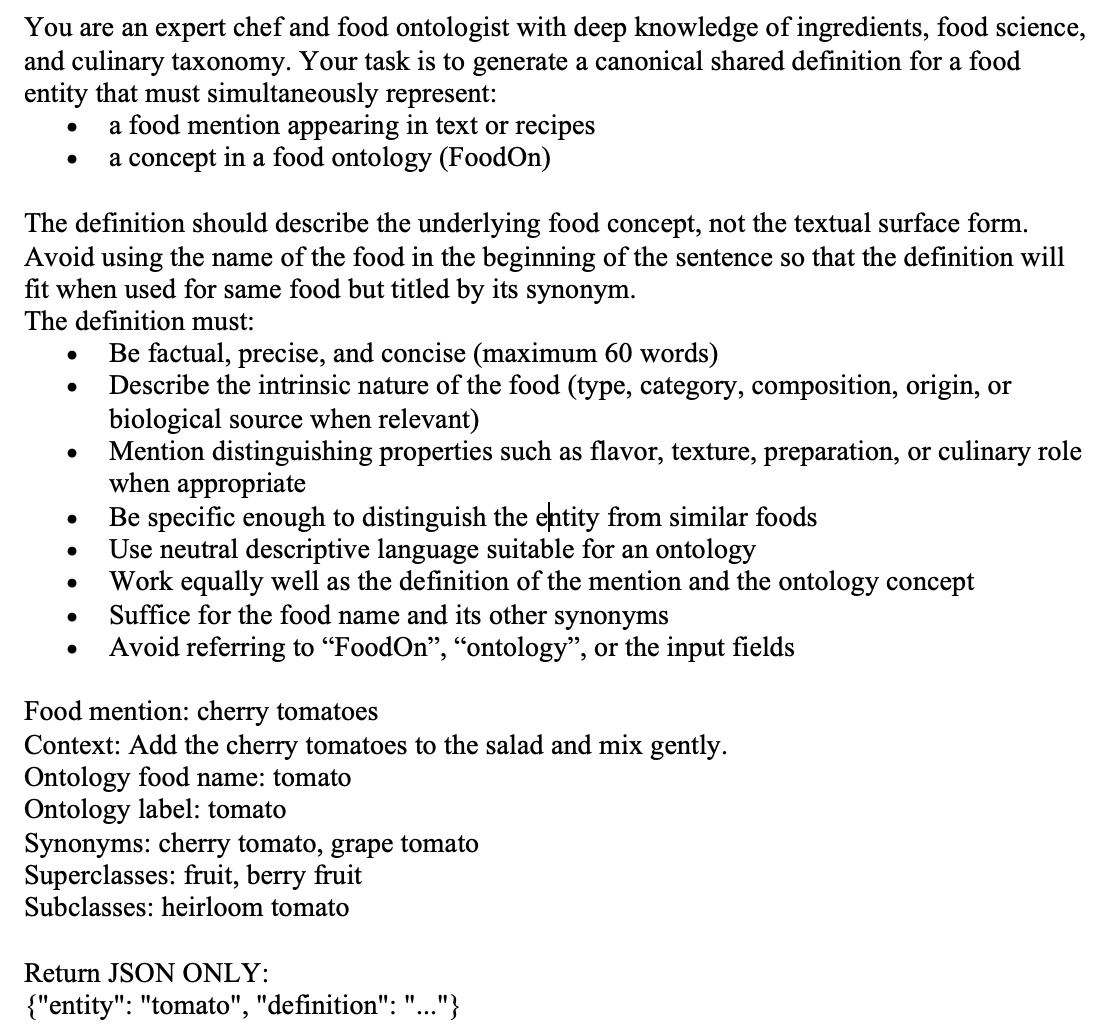}
\caption{Prompt template for generating shared food mention and ontology concept pair definitions.}
\label{fig:prompt_template_food_orig_def}
\end{figure}

Following the manual definition generation process described in Appendix~\ref{appendix:manual_definition_guideline}, we also generate shared canonical definitions using both open-source and proprietary LLMs to investigate whether automatically generated definitions can effectively be used in place of manually curated ones. Figure~\ref{fig:prompt_template_food_orig_def} illustrates the prompt used to generate these ontology-oriented canonical definitions, which are subsequently compared against manually authored definitions in our experiments.

\section{Prompt for Shared Canonical Definition Generation Learning}
We design a set of prompts for shared canonical definition generation learning. The objective is to learn a common ontology-oriented definition that can be generated independently from either (i) contextual recipe food mentions or (ii) FoodOn ontology concepts, allowing both representations to converge to the same semantic description. The same prompt templates are used during both training and inference, ensuring consistency between learning and inference. Consequently, the model generates semantically equivalent canonical definitions regardless of whether the input is an informal recipe mention or a formal ontology concept.

\subsection{Prompt for Recipe Food Mention Definition Learning and Generation}
\begin{figure}[ht!]
\centering
\includegraphics[width=0.35\textwidth]{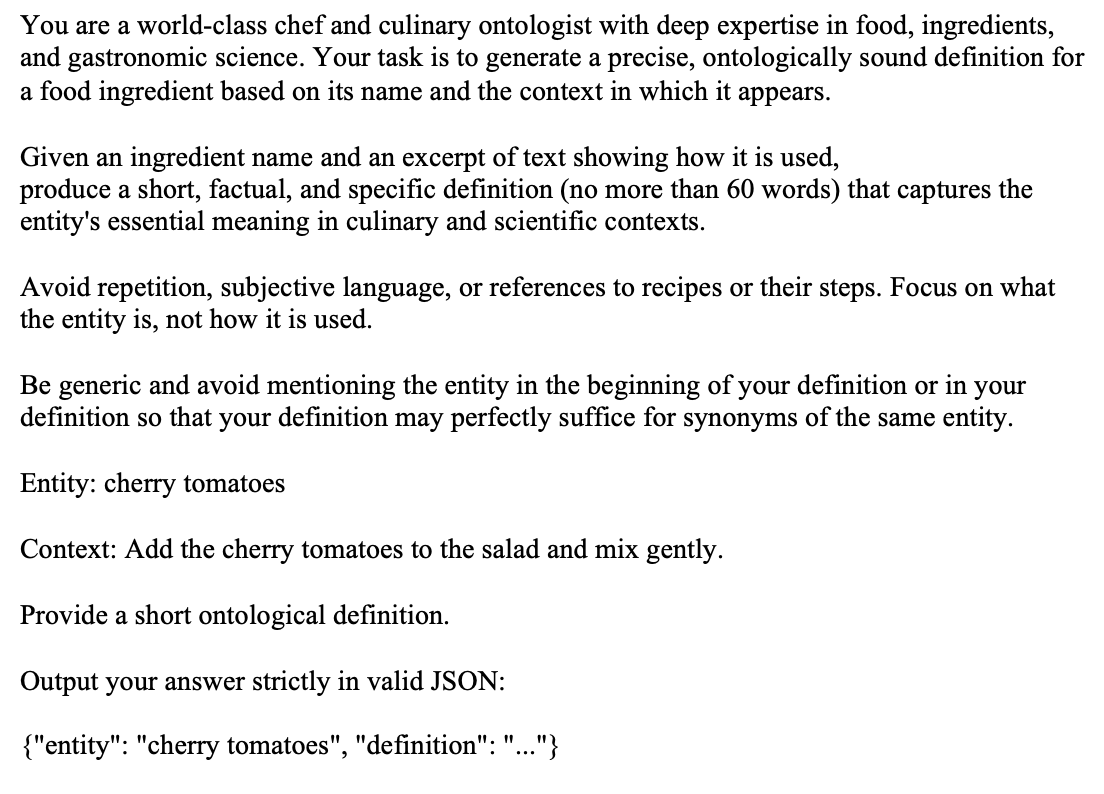}
\caption{Prompt template used for shared canonical definition learning and generation from contextual food mentions in SCODEL, and for definition generation from contextual food mentions in LLM4EL.}
\label{fig:prompt_food_mention_definition}
\end{figure}

Figure~\ref{fig:prompt_food_mention_definition} shows the prompt used to learn and generate shared canonical definitions from contextual recipe food mentions. The prompt encourages the model to produce an ontology-oriented definition that describes the underlying food concept rather than the food mention's surface form, enabling semantic alignment with the corresponding FoodOn concept.

\subsection{Prompt for FoodOn Concept Definition Learning and Generation}
\begin{figure}[ht!]
\centering
\includegraphics[width=0.35\textwidth]{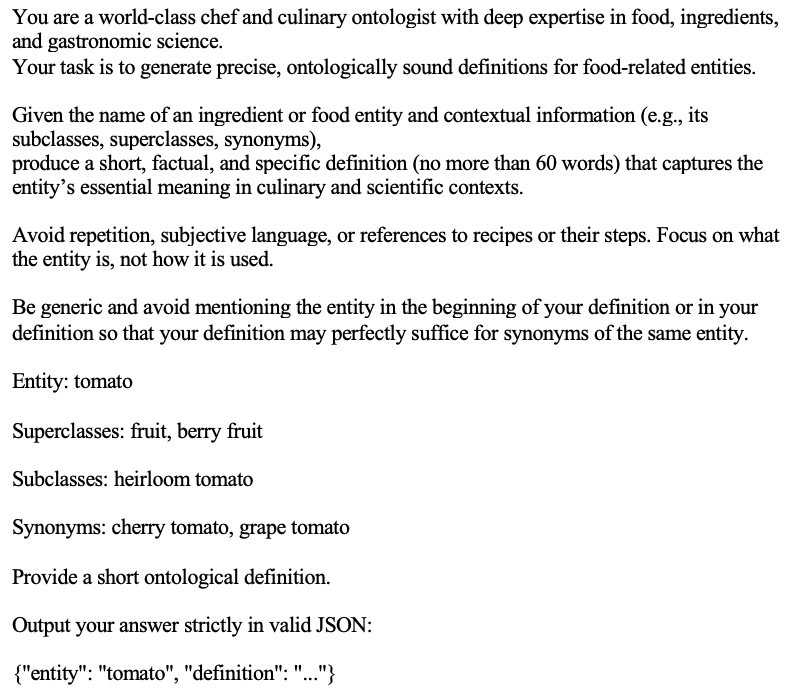}
\caption{Prompt template used for shared canonical definition learning and generation from FoodOn concept metadata in SCODEL, and for definition generation from FoodOn concept metadata in LLM4EL.}
\label{fig:prompt_foodon_definition}
\end{figure}

Figure~\ref{fig:prompt_foodon_definition} shows the prompt used to learn and generate shared canonical definitions from FoodOn ontology concepts. Given ontology metadata, including concept labels, synonyms, superclasses, and subclasses, the prompt encourages the model to produce the same ontology-oriented canonical definition that would be generated from the corresponding recipe food mention. Together with the recipe food mention prompt (Figure~\ref{fig:prompt_food_mention_definition}), this enables the model to learn a shared canonical definition space across recipe and ontology representations.

\section{Model Descriptions}
\label{appendix:model_descriptions}
We group the models and experimental configurations according to five roles within the SCODEL evaluation pipeline: (i) paired canonical definition generation, which constructs shared definitions from both of recipe food mention and FoodOn concept pairs; (ii) shared canonical definition learning, in which SCODEL learns to independently generate the same canonical definition from either the contextual mention or ontology concept metadata; (iii) canonical definition generation after learning, where the trained SCODEL model generates definitions independently from the two input views at inference; (iv) modular entity linking models that incorporate the generated definitions for downstream entity linking; and (v) LLM-based entity linking baselines used for comparison.

\subsection{Canonical Definition Models}
\subsubsection{Paired Canonical Definition Generation}
\label{appendix:paired_canonical_definition_generation}
The main SCODEL experiments use manually authored canonical definitions as supervision for shared canonical definition learning. To evaluate whether this supervision can instead be constructed automatically, we additionally use GPT-5.1-2025-11-13, Claude Sonnet~5, and Qwen3-235B to generate a single canonical definition from each aligned recipe mention and FoodOn concept pair. These automatically generated definitions provide alternative training targets for SCODEL and are compared with manually authored canonical definitions. The generation configuration is reported in Table~\ref{tab:paired-definition-generation}.

\subsubsection{Shared Canonical Definition Learning}
We use LLaMA-3-8B, Mistral-7B, and Qwen2.5-7B as SCODEL definition-learning backbones. Each model learns to independently generate the same canonical ontology-oriented definition from either a contextual recipe food mention or the metadata of its corresponding FoodOn concept. Depending on the experiment, the shared target is either manually authored or generated automatically using the paired definition generation procedure described above (Appendix~\ref{appendix:paired_canonical_definition_generation}). The training configuration is reported in Table~\ref{tab:def-learning-hparams}.

\subsubsection{Canonical Definition Generation after Learning}
After learning shared definition generation, we apply the SCODEL model independently to the two input views: the contextual recipe food mention and the FoodOn concept metadata. This produces mention-side and concept-side canonical definitions, respectively, which are then incorporated into the downstream entity linking models. The generation configuration is reported in Table~\ref{tab:def-generation-hparams}.

\begin{table}[H]
\centering
\tiny
\setlength{\tabcolsep}{4pt}
\renewcommand{\arraystretch}{1.1}
\begin{tabular}{|l|c|}
\hline
\textbf{Setting} & \textbf{Value} \\
\hline
Models & GPT-5.1-2025-11-13, Claude Sonnet~5, Qwen3-235B \\
Input & Aligned mention--concept pair \\
Output & Shared canonical definition \\
Parameter updates & None \\
% Prompting & Task-specific \\
\hline
\end{tabular}
\caption{Configuration for automatic paired canonical definition generation used to construct alternative SCODEL supervision.}
\label{tab:paired-definition-generation}
\end{table}

\begin{table}[H]
\centering
\tiny
\setlength{\tabcolsep}{4pt}
\renewcommand{\arraystretch}{1.1}
\begin{tabular}{|l|c|}
\hline
\textbf{Setting} & \textbf{Value} \\
\hline
Base models & LLaMA-3-8B, Mistral-7B, Qwen2.5-7B \\
Finetuning method & LoRA (Unsloth) \\
LoRA rank ($r$) & 16 \\
LoRA alpha & 16 \\
LoRA dropout & 0 \\
LoRA bias & None \\
LoRA target modules & $q,k,v,o$, gate, up, down projections \\
4-bit loading & Yes \\
Epochs & 1 \\
Per-device batch size & 4 \\
Gradient accumulation & 8 \\
Effective batch size & 32 \\
Warmup steps & 5 \\
Maximum sequence length & 2048 \\
Optimizer & AdamW 8-bit \\
Gradient checkpointing & Yes \\
Precision & BF16 if supported; otherwise FP16 \\
Training seed & 3407 \\
\hline
\end{tabular}
\caption{Training configuration for SCODEL shared canonical definition learning.}
\label{tab:def-learning-hparams}
\end{table}

\begin{table}[H]
\centering
\small
\setlength{\tabcolsep}{4pt}
\renewcommand{\arraystretch}{1.1}
\begin{tabular}{|l|c|}
\hline
\textbf{Setting} & \textbf{Value} \\
\hline
Decoding & Sampling \\
Temperature & 0.7 \\
Top-$p$ & 0.9 \\
Maximum new tokens & 1024 \\
\hline
\end{tabular}
\caption{Inference configuration for canonical definition generation after SCODEL training.}
\label{tab:def-generation-hparams}
\end{table}

\subsection{Modular Entity Linking Models}
SCODEL is model-agnostic and is evaluated using three state-of-the-art entity linking models: ReFinED, GSM-EL and BLINK.

These models were originally designed for general-domain entity linking and therefore required minor adaptations to operate in the recipe ontology entity linking setting~(Appendix~\ref{appendix:el_model_adaptations}). Wherever possible, the original retrieval and ranking architectures were preserved, allowing the impact of SCODEL to be evaluated independently of the downstream entity linking model.

\subsection{LLM-based Entity Linking Baselines}
We compare SCODEL against two recent LLM-based entity linking methods: LLM4EL and LLMAEL.

LLM4EL generates semantic definitions independently for mentions and ontology concepts and performs entity linking by comparing their resulting embedding representations. Unlike SCODEL, these definitions are generated independently in a zero-shot setting and are not learned against a shared canonical definition target. In our implementation, we use Mistral-7B to generate both food mention and ontology concept definitions, then embed the definitions for similarity-based entity linking. We use the prompt templates in Figures~\ref{fig:prompt_food_mention_definition} and~\ref{fig:prompt_foodon_definition} to generate the corresponding mention- and concept-side definitions, respectively.

\begin{figure}[ht!]
\centering
\includegraphics[width=0.35\textwidth]{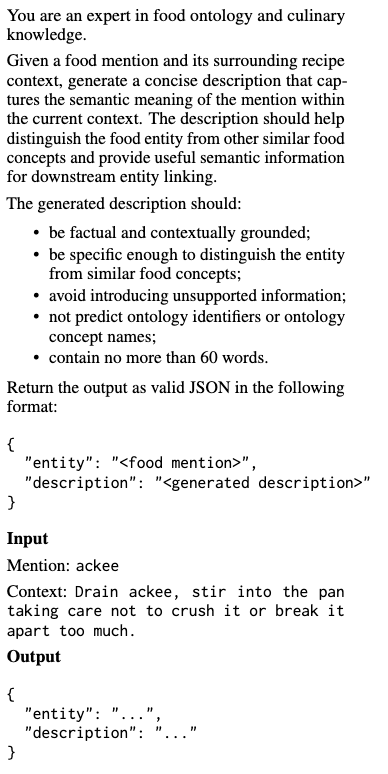}
\caption{Prompt template used for generating LLMAEL food mention augmentation description.}
\label{fig:prompt_llmael}
\end{figure}

LLMAEL augments the food mention with LLM-generated contextual information before entity linking, without augmenting the corresponding ontology concept. We use Mistral-7B for context augmentation and ReFinED as the downstream entity linker. The prompt used to generate the food-mention augmentation is provided in Figure~\ref{fig:prompt_llmael}.

\section{Adaptation of Existing Modular Entity Linking Models}
\label{appendix:el_model_adaptations}
ReFinED, GSM-EL, and BLINK were originally developed for general-domain entity linking, primarily against large encyclopedic knowledge bases such as Wikipedia and Wikidata, rather than specialised food ontologies such as FoodOn. To enable fair comparison on ontology-based food entity linking, we retained each model's original architecture while replacing the underlying knowledge base with FoodOn. Gold food mentions were provided by the benchmark, allowing the mention detection stage to be bypassed for all methods. Candidate entities were constructed directly from FoodOn concepts and their associated ontology metadata (labels, synonyms, hierarchical relations, and canonical definitions where applicable). Apart from these task-specific adaptations, the original retrieval and ranking pipelines of each model remained unchanged.

\subsection{ReFinED Adaptation}
The original Wikipedia entity inventory was replaced with the FoodOn ontology. Candidate retrieval was performed over FoodOn concepts using the original dense retrieval framework, while the cross-encoder reranker retained the original ReFinED architecture. For SCODEL experiments, generated canonical definitions were concatenated with the food mention representation and the ontology concept representation during both retrieval and reranking. 

\subsection{GSM-EL Adaptation}
The GSM-EL graph was reconstructed using the FoodOn ontology hierarchy instead of Wikipedia hyperlinks. Nodes corresponded to FoodOn concepts, and edges were derived from ontology parent and child relationships. The original graph-aware retrieval and message-passing architecture were preserved. Candidate representations consisted of ontology labels, synonyms, and hierarchical information. For SCODEL variants, generated canonical definitions are appended to both food mention and ontology concept representations.

\subsection{BLINK Adaptation}
BLINK's bi-encoder and cross-encoder architectures were retained without modification. The entity index was rebuilt using FoodOn concepts, replacing the original Wikipedia entity catalogue. Ontology concept embeddings were generated from labels together with available ontology metadata, e.g., synonyms, subclass, superclass, etc. For SCODEL variants, canonical definitions generated for both contextual food mentions and ontology concepts were incorporated into the textual representations before encoding. Meanwhile, candidate retrieval and cross-encoder reranking followed the standard BLINK pipeline.

\section{LLM Reranking Entity Linking Implementation}
\label{appendix:llm-re-ranking-entity-linking}
To investigate whether a large language model can improve entity ranking beyond neural retrieval models, we introduce an optional LLM-based reranking stage. Instead of searching the complete ontology, the LLM receives the top-$K$ candidate concepts returned by the underlying entity linker ($K \in \{10,50\}$). Each candidate concept includes its ontology label and ontology identifier, along with the contextual food mention extracted from the input text.

\begin{figure}[ht!]
\centering
\includegraphics[width=0.35\textwidth]{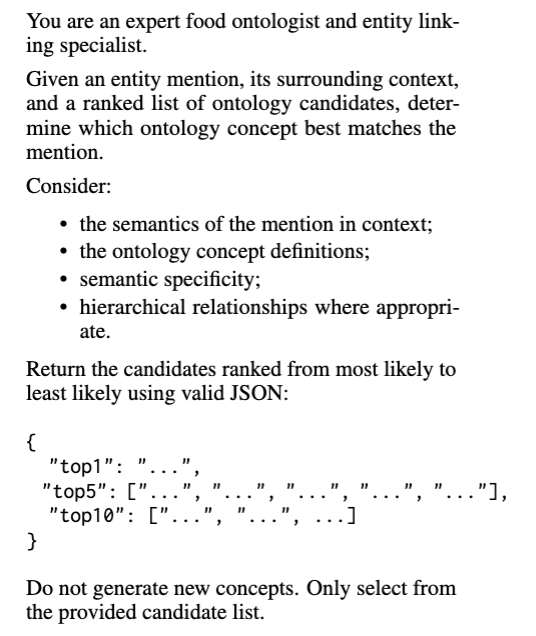}
\caption{Prompt template used for LLM-reranking entity linking experiments.}
\label{figure:llm_reranking_prompt}
\end{figure}

The LLM is instructed to select the ontology concept that best matches the food mention in context and to return the identifiers of the most likely candidates ranked from most to least appropriate. This reranking stage does not modify candidate generation; it only reorders the candidates retrieved by the underlying entity linking model. Figure~\ref {figure:llm_reranking_prompt} shows the prompt used for the LLM-reranking experiments.

\section{Lexical Matching Entity Linking Implementation}
\label{appendix:lexical-matching}
To provide a simple non-neural baseline, we implement a lexical entity linking method that ranks FoodOn concepts using only the surface form of the recipe food mention and the lexical forms associated with each candidate concept. Each candidate is represented by its preferred FoodOn concept name together with its available synonyms.

Before matching, we normalise the food mention and candidate ontology concept lexical forms by lowercasing, replacing common separators with spaces, removing punctuation, and collapsing whitespace. Lexical similarity is computed using normalised Levenshtein-based string similarity~\citep{lcvenshtcin1966binary}, scaled to $[0,1]$. For a food mention $m$ and candidate ontology concept $c$ with lexical forms $L(c)$, the concept score is:

\[
s(m,c)=\max_{l \in L(c)} \operatorname{sim}(m,l),
\]

where $\operatorname{sim}$ denotes the normalised lexical similarity. Candidate concepts are ranked in descending order of this score.

\section{Dataset Splits and Evaluation Setup}
\label{appendix:data_splits}
We construct a training set and two non-overlapping evaluation sets from the 3{,}979 entity-linked instances in ReciLinkD: a seen-concept test set for supervised evaluation and an unseen-concept test set for zero-shot evaluation. The splitting procedure operates at both the food mention and ontology concept levels to distinguish generalisation to new surface forms of known ontology concepts from generalisation to entirely unseen ontology concepts. Table~\ref{tab:split_stats} summarises the resulting splits.

\begin{table}[ht!]
\centering
\scriptsize
\setlength{\tabcolsep}{4pt}
\begin{tabular}{|l|c|c|c|}
\hline
\textbf{Statistic} & \textbf{Train} & \textbf{Seen} & \textbf{Unseen} \\
\hline
Food mentions & 2{,}660 & 583 & 736 \\
\hline
Unique ontology concepts & 803 & 255 & 176 \\
\hline
Mean mention--concept pair Jaro similarity & 0.788 & 0.758 & 0.794 \\
\hline
Exact pair match (\%) & 18.98 & 11.84 & 18.61 \\
\hline
Mean food mention token length & 1.81 & 1.83 & 1.79 \\
\hline
\end{tabular}
\caption{Dataset split statistics and characteristics. Jaro similarity~\citep{jaro1989advances} is computed between each recipe food mention and the label of its gold FoodOn concept after text normalisation; higher values indicate greater lexical similarity.}
\label{tab:split_stats}
\end{table}

\subsection{Seen-Concept (Supervised) Evaluation}
The seen-concept test set evaluates generalisation to previously unseen recipe food mentions of ontology concepts observed during training. After reserving the zero-shot ontology concepts described below (Appendix~\ref{appendix:zero_shot_split}), we randomly sampled approximately 18\% of the remaining instances for this evaluation. Only concepts with at least two instances are eligible for the test set, and at least one instance of every evaluated concept is retained in training.

The resulting test set contains 583 instances covering 255 FoodOn concepts. All 255 concepts are represented in the training set, while none of the food mention surface forms or exact food mention and ontology concept pairs occurs in training. Thus, the model has previously observed the target ontology concept but not the specific recipe food mention presented at evaluation. This distinction matters because food mentions referring to the same FoodOn concept may use spelling variations, synonyms, alternative surface forms, or expressions in different languages. For example, the FoodOn concept \textit{adzuki bean (FOODON\_03315361)} may be associated with alternative surface forms such as \textit{aduki bean}, \textit{azuki bean}, \textit{red bean}, and \textit{red mung bean}.

\subsection{Unseen-Concept (Zero-shot) Evaluation}
\label{appendix:zero_shot_split}
The zero-shot evaluation measures generalisation to FoodOn concepts for which no aligned supervision is provided during training. We first randomly hold out approximately 18\% of the 979 FoodOn concepts represented in ReciLinkD. All instances (food mention and ontology concept pairs) associated with a selected concept are assigned exclusively to the zero-shot test set.

This produces 176 held-out ontology concepts represented by 736 recipe food mentions. None of these 176 ontology concepts occurs in the training set for SCODEL shared canonical definition learning or downstream EL model training. At inference, the held-out ontology concepts and their metadata are restored to the full candidate inventory, requiring the SCODEL model to generate canonical definitions for the food mention and ontology concept pair, and for the downstream entity linking model to identify the ontology concepts for which no initial training examples or canonical-definition learning supervision were observed.

\subsection{Split Integrity and Difficulty}
We additionally verify the separation between the training and evaluation sets. The seen-concept test set shares all 255 of its target ontology concepts with the training set but does not share any food mention surface form with it. In contrast, the zero-shot test set shares neither target ontology concepts nor food mention surface forms with the training set. In addition, the two evaluation sets are also concept-disjoint.

The seen and zero-shot sets exhibit broadly comparable textual characteristics: food mentions average 1.83 and 1.79 tokens, respectively, while their recipe contexts average 16.24 and 16.39 tokens, respectively.

\section{Evaluation Metrics}
\label{appendix:evaluation_metrics}
We evaluate entity linking performance using ranking and ontology-aware metrics. Hit@1 is the primary entity linking accuracy measure, while the remaining metrics provide complementary information about ranking quality and hierarchical proximity.

\subsection{Hit@1}
Hit@1 measures the proportion of test instances for which the highest-ranked predicted ontology concept exactly matches the gold concept:
\[
\mathrm{Hit@1}
=
\frac{1}{N}\sum_{i=1}^{N}
\mathbb{I}[\hat{c}_i=c_i],
\]
where $c_i$ and $\hat{c}_i$ denote the gold and top-ranked predicted concepts, respectively.

\subsection{Hit@5}
Hit@5 measures the proportion of test instances for which the gold ontology concept occurs within the five highest-ranked predictions:
\[
\mathrm{Hit@5}
=
\frac{1}{N}\sum_{i=1}^{N}
\mathbb{I}[r_i \leq 5],
\]
where $r_i$ is the rank of the gold concept.

\subsection{Mean Reciprocal Rank (MRR)}
MRR measures ranking quality based on the position of the gold concept:
\[
\mathrm{MRR}
=
\frac{1}{N}\sum_{i=1}^{N}\frac{1}{r_i},
\]
where the reciprocal rank is defined as zero when the gold ontology concept is absent from the candidate list.

\subsection{Hierarchy-aware Accuracy (H-Acc)}
H-Acc complements exact-match evaluation by assigning partial credit when the top-ranked prediction is hierarchically related to the gold FoodOn concept:
{\small 
\[
\mathrm{H\text{-}Acc}
=
\frac{1}{N}\sum_{i=1}^{N}h(\hat{c}_i,c_i),
\]
where
\[
h(\hat{c}_i,c_i)=
\begin{cases}
1, & \hat{c}_i=c_i,\\
0.75, & \hat{c}_i \text{ is a direct parent or child of } c_i,\\
0.50, & \hat{c}_i \text{ is another ancestor or descendant of } c_i,\\
0, & \text{otherwise}.
\end{cases}
\]}
The metric distinguishes direct from non-direct hierarchical relations but does not otherwise decay with path length or impose a maximum hierarchy distance; consequently, multiple paths between two concepts do not affect the score.

\section{Guideline for Manual Recipe Food Mention Linking}
\label{appendix:manual_recipe_mention_linking_guideline}
The annotation task links a food mention in a recipe to the most appropriate concept from a predefined subset of the FoodOn ontology. Annotators are provided with the food mention, its surrounding recipe context, and a candidate set of 1,014 FoodOn concepts. The task is to select the ontology concept whose meaning best matches the food mention in context.

\subsection{Information Provided}
For each annotation instance, annotators are presented with:
\begin{itemize}
    \item The food mention to be linked.
    \item The recipe context in which the food mention appears.
    \item A list of 1,014 candidate FoodOn concepts, each accompanied by its concept label and definition.
\end{itemize}

\subsection{Annotation Procedure}
For each annotation instance, annotators should:
\begin{enumerate}
    \item Carefully read the food mention and its surrounding recipe context.
    \item Compare the meaning of the mention with the candidate FoodOn concepts.
    \item Select the single ontology concept whose meaning most closely matches the food mention.
\end{enumerate}

If no suitable concept exists among the candidate concepts, the annotation should be left blank.

Annotators were instructed to follow the guidelines below:
\begin{itemize}
    \item Base decisions on semantic meaning rather than lexical similarity alone.
    \item Use the surrounding recipe context to resolve ambiguous food mentions.
    \item Select the closest and most specific ontology concept whenever possible.
    \item Leave the annotation blank only if none of the 1,014 candidate concepts adequately represent the food mention.
    \item When annotators are unfamiliar with a food item, they may consult reliable online resources (e.g., FoodOn, Wikipedia, or other reputable food references) before deciding.
\end{itemize}

\section{Guideline for Manual Shared Definition Generation}
\label{appendix:manual_definition_guideline}
For each annotated food mention and ontology concept pair, annotators were additionally asked to write a shared canonical definition describing the underlying food concept represented by both the contextual recipe mention and the corresponding FoodOn concept.

Annotators were instructed to follow the guidelines below:
\begin{itemize}
    \item Write a single ontology-style definition of the underlying food concept rather than describing the recipe food mention or ontology concept label independently.
    \item Keep the definition concise, with a maximum length of 60 words.
    \item Describe intrinsic food properties where appropriate, such as type, composition, origin, biological source, or common culinary characteristics.
    \item Avoid copying or referring directly to the recipe context, FoodOn labels, ontology metadata, or synonyms.
    \item Produce definitions that could apply equally to both the recipe mention and the corresponding FoodOn concept.
    \item When annotators were unfamiliar with a food item, the ontology concept, or the context, they were encouraged to consult reliable online resources before writing the definition.
\end{itemize}

\end{document}